\documentclass{article} %
\usepackage{iclr2027_conference,times}

\usepackage{amsmath,amsfonts,bm}

\newcommand{\figleft}{{\em (Left)}}
\newcommand{\figcenter}{{\em (Center)}}
\newcommand{\figright}{{\em (Right)}}

\def\eqref#1{equation~\ref{#1}}

\def\1{\bm{1}}

\DeclareMathAlphabet{\mathsfit}{\encodingdefault}{\sfdefault}{m}{sl}
\SetMathAlphabet{\mathsfit}{bold}{\encodingdefault}{\sfdefault}{bx}{n}

\def\gA{{\mathcal{A}}}

\def\gD{{\mathcal{D}}}

\def\gL{{\mathcal{L}}}

\def\gN{{\mathcal{N}}}

\def\gS{{\mathcal{S}}}
\def\gT{{\mathcal{T}}}

\def\gX{{\mathcal{X}}}

\def\gZ{{\mathcal{Z}}}

\def\sR{{\mathbb{R}}}

\newcommand{\E}{\mathbb{E}}

\newcommand{\R}{\mathbb{R}}

\DeclareMathOperator*{\argmax}{arg\,max}
\DeclareMathOperator*{\argmin}{arg\,min}

\usepackage[utf8]{inputenc} %
\usepackage[T1]{fontenc}    %
\usepackage[colorlinks]{hyperref}       %
\usepackage[capitalise]{cleveref}
\usepackage{url}            %
\usepackage{booktabs}       %
\usepackage{amsfonts}       %
\usepackage{microtype}      %
\usepackage{xcolor}         %
\usepackage{bbm}
\usepackage{amssymb}
\usepackage{amsthm}
\usepackage{amsmath}
\usepackage{mathtools}
\usepackage{enumitem}
\usepackage{algorithm}
\usepackage{algpseudocode}

\usepackage[T1]{fontenc}
\usepackage{booktabs}
\usepackage{makecell}
\usepackage{pifont}
\usepackage{graphicx}
\usepackage{subcaption}
\usepackage{tcolorbox}
\usepackage{tikz}
\usetikzlibrary{patterns}

\AddToHook{cmd/ttfamily/after}{\frenchspacing}
\newcommand{\de}{\mathrm{d}}

\newcommand{\cmark}{\ding{51}}
\newcommand{\xmark}{\ding{55}}

\newcommand{\methodhead}[2]{%
    \makecell[c]{%
        \texttt{#1}\\[-1pt]
        {\tiny\citeauthor{#2}}\\[-2pt]
        {\tiny\citeyearpar{#2}}%
    }%
}

\newcommand{\ourmethodhead}[1]{%
    \makecell[c]{%
        \texttt{#1}\\[-1pt]
        {\tiny\texttt{(Ours)}}\\[-2pt]
        {\tiny\phantom{(2026)}}%
    }%
}

\newtcolorbox{greenbox}[1][]{
    colback=mygreen!5!white,
    colframe=mygreen,
    colbacktitle=mygreen,
    coltitle=white,
    fonttitle=\bfseries,
    boxrule=0.7pt,
    arc=1.5mm,
    left=2mm,
    right=2mm,
    top=1mm,
    bottom=1mm,
    #1
}

\newenvironment{footalign}
  {\footnotesize\align}
  {\endalign\noindent}
\newenvironment{footalign*}
  {\footnotesize\csname align*\endcsname}
  {\csname endalign*\endcsname\noindent}

\newenvironment{footgather*}
  {\footnotesize\csname gather*\endcsname}
  {\csname endgather*\endcsname\noindent}

\DeclarePairedDelimiter\norm{\lVert}{\rVert}
\DeclarePairedDelimiter\abs{\lvert}{\rvert}

\makeatletter
\newcommand{\newreptheorem}[2]{\newtheorem*{rep@#1}{\rep@title}\newenvironment{rep#1}[1]{\def\rep@title{#2 \ref*{##1}}\begin{rep@#1}}{\end{rep@#1}}}
\makeatother

\newreptheorem{theorem}{Theorem}
\newtheorem{proposition}{Proposition}
\newreptheorem{proposition}{Proposition}
\newtheorem{lemma}{Lemma}
\newreptheorem{lemma}{Lemma}
\newtheorem{remark}{Remark}
\newreptheorem{remark}{Remark}

\newreptheorem{corollary}{Corollary}

\newreptheorem{definition}{Definition}

\definecolor{mygreen}{HTML}{2ca02c}
\definecolor{mygreendark}{HTML}{1C661C}
\definecolor{myblue}{HTML}{1f77b4}
\definecolor{myorange}{HTML}{ff7f0e}
\definecolor{mypink}{HTML}{e377c2}
\definecolor{mypurple}{HTML}{9467bd}
\definecolor{mygray}{HTML}{878787}
\hypersetup{urlcolor=mygreen, citecolor=mygreen, linkcolor=mygreen}

\newcommand{\TwoLineComment}[2][mygray]{%
    \unskip\hfill\raisebox{0.5\baselineskip}[0pt][0pt]{%
        \textcolor{#1}{$\triangleleft$ #2}}%
}

\DeclareRobustCommand{\distRLlegend}{%
  \tikz[baseline={([yshift=-0.55ex]current bounding box.center)}]{%
    \draw[pattern=north east lines, line width=0.3pt]
      (0,0) rectangle (0.35,0.18);
  }%
}

\title{Value Forcing: Aligning Return Distributions at All Noise Levels}

\author{%
    \begin{minipage}{\textwidth}
    \centering
    \vspace{0.75em}
    Chongyi Zheng$^{1}$ \quad Zhiyuan Zhou$^{2}$ \quad Qiyang Li$^{2}$ \\
    \vspace{0.3em}
    Benjamin Eysenbach$^1$ \quad Sergey Levine$^2$ \\
    \vspace{0.3em}
    \normalfont $^1$Princeton University \qquad $^2$University of California, Berkeley \\
    \vspace{0.5em}
    \texttt{chongyiz@princeton.edu}
    \end{minipage}
}

\newcommand{\methodname}[0]{VF}
\newcommand{\methodnamei}[0]{IVF}
\newcommand{\methodexplain}{Value Forcing}

\iclrfinalcopy %
\begin{document}

\maketitle

\vspace{-1em}
\begin{abstract}
\vspace{-0.5em}
Learning accurate value functions is central to reinforcement learning (RL), yet remains challenging because supervision is limited and indirect: ground-truth values are unavailable, and each transition in the data only provides an immediate reward that can be used to construct bootstrapped targets as supervision. This challenge is particularly pronounced when data is limited or reward signals are sparse, as is common in real-world applications such as robotics. 
We argue that the key to accurate value learning in this \textit{limited} data regime is extracting richer learning signals from the available data. In this paper, we introduce \textbf{\methodexplain{}}, a new approach that uses flow matching to estimate return distributions and enables accurate value learning from limited data. Our key insight is that we can extract richer learning signals by enforcing Bellman consistency at every noise level. Empirically, \methodexplain{} improves the accuracy of estimating the return distribution, and prevents the distribution from collapsing to a single value, as occurs with prior methods. We instantiate \methodexplain{} with two practical value-learning algorithms: \methodname{}, an actor-critic method, and \methodnamei{}, an implicit value-learning method. We evaluate both algorithms across $26$ long-horizon tasks in the offline RL setting and find that they outperform prior value-learning methods, with particularly large gains when data is limited. Our project website is available at \url{https://chongyi-zheng.github.io/value-forcing}.
\end{abstract}

\section{Introduction}

\begin{figure}[h]
    \centering
    \includegraphics[width=\linewidth]{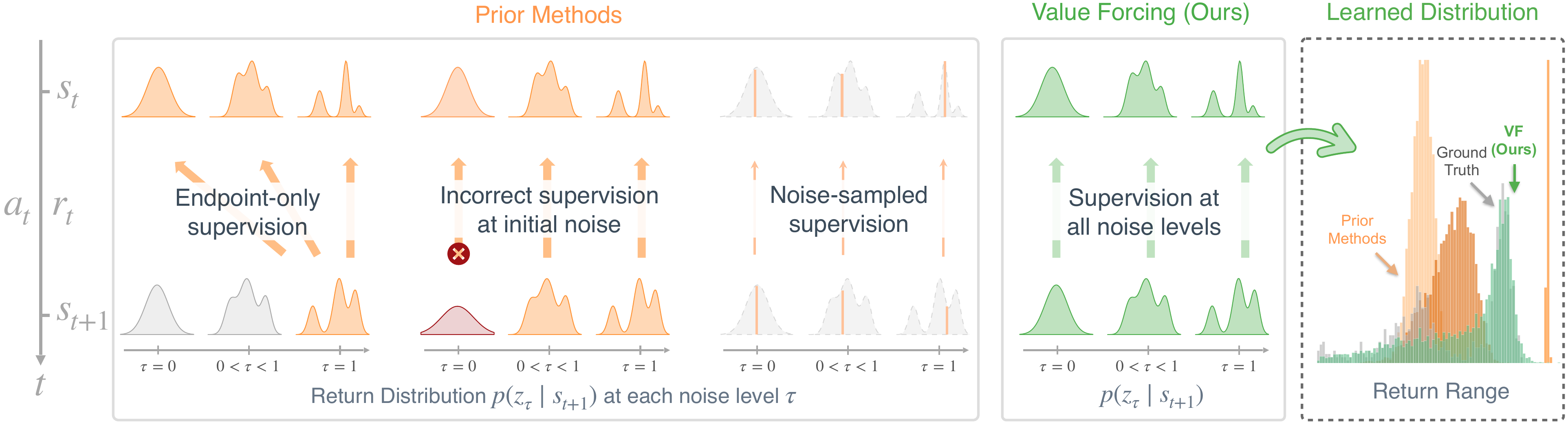}
    \caption{
    We introduce \textbf{\methodexplain{}}, a new approach to learning distributional value functions with flow matching that supervises the learned return distribution at every flow noise level through Bellman consistency.
    Unlike prior methods that provide supervision only at selected noise levels or use inaccurate targets at others, \methodexplain{} provides substantially richer learning signals and enables much more accurate estimation of the return distribution, preventing it from collapsing around a single value. This in turn leads to better policy performance, especially in the limited-data regime.
    }
    \label{fig:teaser}
\end{figure}

Value estimation lies at the heart of reinforcement learning (RL): value functions estimate the expected return of following a certain course of actions, guiding agents to select better actions that improve long-term returns. However, accurately estimating the values is hard, especially with \textit{limited} amounts of data or if the reward feedback is sparse. Supervision for long-term value is limited and indirect: each observed transition provides only an immediate reward, but no ground-truth expected return. Temporal-difference (TD) methods construct prediction targets using their own value estimates, which could be inaccurate and bias learning. With limited data, there is little room to correct these errors. 
However, such settings are common in real-world applications such as robotics. This raises the question: \textit{is there a way to extract richer learning signals from limited available transitions?}

Many widely used RL methods~\citep{mnih2015human, schulman2017proximal, haarnoja2018soft} learn scalar value functions that summarize the future outcome of a policy. This scheme has limited expressivity and may give a sparse supervision signal when used as the target for TD learning.
Distributional RL instead models the full return distribution and can provide richer learning signals when the whole return distribution is used as the TD target~\citep{wang2023benefits, rowland2023statistical}. For example, an action that produces returns of $-1$ and $+1$ with equal probability has the same expected return as another action that always produces $0$ return, yet their distributions differ. Learning to distinguish these two actions and their outcomes can enable more effective use of the available data. 

Recent work has begun to use modern generative models such as flow matching to represent the return distribution. These methods learn to map Gaussian noise to the return distribution through an iterative denoising process. Beyond capturing complex return distributions, flow matching also creates an opportunity to provide even richer supervision by learning at different noise levels~\citep{farebrother2025temporal, chen2024diffusion}. 
However, prior flow-based return estimators typically only provide supervision using the endpoint of the flow path~\citep{chen2025unleashing, zhong2025flowcritic, groom2026quantile, agrawalla2026floq}, missing the opportunity to supervise all other noise levels. While others have explored applying Bellman consistency at all noise levels, they either suffer from enforcing an incorrect consistency at the initial flow step~\citep{dong2025value} or only apply the consistency along the flow paths starting from the same noise~\citep{xu2026path} (Fig.~\ref{fig:teaser} \figcenter{}).
Empirically, we find that sparse supervision or incorrect dense supervision can lead to inaccurate modeling of the return distribution, and often results in the learned return distribution collapsing toward a point mass (see Fig.~\ref{fig:ret-hist}), thereby undermining the effectiveness of distributional value learning.

In this paper, we introduce \methodexplain{}, a new distributional value learning approach that enforces Bellman consistency at every noise level (see illustration in Fig.~\ref{fig:teaser}). At each noise level, \methodexplain{} uses the next-state marginal return distribution to supervise the current-state marginal return distribution.
Unlike the standard Bellman operator, we introduce a new flow Bellman operator that includes a Gaussian correction.
We show that \methodexplain{} has a unique fixed point at every noise level, preserving the initial Gaussian noise and recovering the distributional Bellman equation in the denoised space.
Empirically, we find that \methodexplain{} can model the return distribution much more accurately compared to the prior methods and does not suffer from distribution collapse (Fig.~\ref{fig:teaser}).

We instantiate two practical value-learning algorithms with \methodexplain{}: \methodname{}, an actor-critic algorithm, and \methodnamei{}, an implicit method via expectile-weighted flow matching. Theoretically, we show that \methodnamei{} approaches the optimal value within actions presented in the dataset as the expectile increases. Empirically, we evaluate both \methodname{} and \methodnamei{} across $26$ long-horizon tasks, $20$ from OGBench and $6$ from OCBench, and show that they improve return estimation and policy performance over prior flow-based value learning methods, with particularly strong gains when data is limited.

\section{Related Work}

\textbf{Offline RL} aims to learn a reward-maximizing policy from a fixed offline dataset.
The core challenge in this setting is to learn a value function while preventing erroneous extrapolation outside the distribution of the dataset. Previous works have proposed strategies based on applying conservative penalties for out-of-distribution samples~\citep{kumar2020conservative, nakamoto2024cal, bai2022pessimistic, chae2026flow}, constraining action selection in Bellman backups to the dataset support~\citep{fujimoto2019off, kumar2019stabilizing, singh2022offline}, or not querying samples outside the offline dataset at all~\citep{kostrikov2022offline, hansen2023idql, xu2023offline}.
While most previous works focus on training an accurate value function with the entire offline dataset, we focus on the setting with \textit{extremely limited} offline data.
Most similar to our focus is the work of~\citet{xu2023offline, park2026pretraining}, which also compares different value learning algorithms on subsets of the entire offline datasets. However, these methods learn a scalar return value, while we aim to learn the whole distribution of the return value using flow matching.
This offers denser supervision during TD updates, and in practice leads to better performance in the limited dataset setting. We find that our method matches the state of the art on the full dataset, but outperforms it by a large margin when data is more limited.

\textbf{Distributional and generative value functions.}
Distributional value functions aim to learn the entire distribution of the discounted cumulative return instead of a single scalar that estimates its expected value~\citep{bellemare2017distributional, bellemare2023distributional}. 
Early methods model the value distribution as a discrete categorical distribution~\citep{bellemare2017distributional, wang2026learning,dabney2018distributional} or continuous distributions with restrictive modeling assumptions~\citep{ma2025dsac, dabney2018implicit}.
Recent methods have leveraged diffusion and flow matching to represent complex return distributions~\citep{dong2025value, zhong2025flowcritic, chen2025unleashing, espinosa2025expressive} 
and their expected values~\citep{agrawalla2026floq}.
Our method falls into the former category, where methods differ primarily in where the Bellman consistency is enforced in the denoising path.
DFC~\citep{chen2025unleashing}, FlowCritic~\citep{zhong2025flowcritic}, FlowIQN~\citep{groom2026quantile}, and Dual-Flow RL~\citep{li2026dual} enforce the Bellman consistency at the endpoint of the flow, where the Bellman target is computed in the denoised space and used as a flow-matching target for all noise levels. FlowIQN further introduces a quantile-aligned coupling between initial noises and final samples. 
\citet{dong2025value},~\citet{espinosa2025expressive}, and~\citet{xu2026path} introduce stronger constraints along the flow path: Value Flows~\citep{dong2025value} and EVOR~\citep{espinosa2025expressive} use the velocity field directly as the TD bootstrapping target, but na\"ively apply the clean Bellman equation at all noise levels; 
PCBF~\citep{xu2026path} couples two flow paths using the same initial noise and applies the Bellman operator at all noise levels along the coupled paths, but the marginal flow paths lack Bellman consistency because of the coupling. In contrast, our method enforces the Bellman consistency at all noise levels. We achieve this by defining a flow Bellman operator whose fixed point aligns the return distributions at every noise level. We find that this results in much more accurate return estimation, especially in the low-data regime. 
See the Appendix Table~\ref{tab:flow-bellman-comparison} for a summary of these differences.

\section{Preliminaries}
\label{sec:prelim}

\textbf{Markov decision process and offline RL.} 
We consider an infinite-horizon Markov decision process
$\mathcal{M} = (\gS, \gA, p, r, \rho, \gamma)$, where $\gS$ and
$\gA$ are the state and action spaces,
$p: \gS \times \gA \to \Delta(\gS)$ is the transition distribution,
$r: \gS \times \gA \to [r_{\text{min}}, r_{\text{max}}]$ is the \emph{deterministic} bounded reward function,
$\rho \in \Delta(\gS)$ is the initial state distribution, and $\gamma \in [0, 1)$ is the discount factor.
Here, $\Delta(\cdot)$ denotes the set of all possible probability distributions. Throughout our discussions, we will use $t \in \mathbb{N}$ to denote the time step for RL and use $\tau \in [0, 1]$ to denote the noise level for flow-matching. We focus on offline RL, where the agent learns from a fixed dataset $D = \{(s, a, r, s')\}$ collected by a behavior policy.
The goal is to learn a policy $\pi: \gS \to \Delta(\gA)$ that maximizes the expected
discounted return: $\E_{\pi} \left[ \sum_{t = 0}^{\infty} \gamma^t r(s_t, a_t) \right]$.

\textbf{Distributional RL.} Instead of modeling the expected discounted return, distributional
RL~\citep{bellemare2017distributional, dabney2018implicit} models the full return distribution. For a policy $\pi$, the return random variable is defined as $Z^{\pi}(s, a) = \sum_{t = 0}^{\infty} \gamma^t r(S_t, A_t)$, where $(S_t, A_t)$ are state and action random variables at time $t$. We use $\gZ$ to denote the set of return distributions with uniformly bounded moments~\citep{bellemare2017distributional} for all possible policies in this MDP. Taking the expectation of $Z^{\pi}(s, a)$ recovers the Q function: $Q^{\pi}(s, a) = \E_{\pi} \left[ Z^{\pi}(s, a) \mid S_0 = s, A_0 = a \right]$. 
Prior work~\citep{bellemare2017distributional} has shown that $Z^{\pi}(s, a)$ satisfies the distributional Bellman equation
\begin{align}
    Z^{\pi}(s, a) \overset{d}{=} \gT^{\pi} Z^{\pi}(s, a) \overset{d}{=} r(s, a) + \gamma Z^{\pi}(S', A'),
    \label{eq:dist-bellman-eq}
\end{align}
where $(S', A')$ are random variables of the state and action at the next time step, $\overset{d}{=}$ indicates equality in distribution, and $\gT^{\pi}$ is the distributional Bellman operator. This operator is a $\gamma$-contraction under the maximum Wasserstein metric with the unique fixed point $Z^{\pi}(s, a)$. Taking expectations of Eq.~\ref{eq:dist-bellman-eq} recovers the standard Bellman equations for the Q function.

\textbf{Flow matching.}
Flow matching~\citep{lipman2022flow, flow_albergo2023, liu2022flow} is a generative modeling framework that learns a continuous velocity field to transport samples from a noise distribution to a target distribution. Let $u(x^\tau \mid \tau): \sR^d \times [0,1] \rightarrow \sR^d$ denote a velocity field that depends on noise levels. The corresponding flow path $\phi(x^0 \mid \tau): \sR^d \times [0,1] \rightarrow \sR^d$ is defined as the solution to the following ordinary differential equation (ODE)~\citep{smooth_lee2002}
\begin{footalign}
    \frac{\de}{\de\tau}
    \phi(x^0\mid\tau)
    &=
    u \left(
        \phi(x^0\mid\tau)
        \mid\tau
    \right),
    &
    \phi(x^0 \mid 0)
    &=
    x^0.
    \label{eq:flow-ode}
\end{footalign}
Starting from an initial noise distribution (e.g., the standard Gaussian $\gN(0, I_d)$), we learn the flow to transport samples to match the target distribution by minimizing the conditional flow-matching objective~\citep{lipman2022flow}:
\begin{align}
    \E \left[
        \left\|v(x^\tau\mid\tau)-(x^1-x^0)\right\|_2^2
    \right],
    \label{eq:standard-flow-matching-loss}
\end{align}
where $x^0 \sim \gN(0, I_d)$ is a noise sample, $x^1$ is a sample from the target distribution, and $x^\tau = (1 - \tau) x^0 + \tau x^1$ is the linear interpolation at noise level $\tau\sim\textsc{Unif}([0, 1])$. Prior work~\citep{liu2022flow, albergo2025stochastic} has shown that this loss is related to the noise-level differentiation of a stochastic process:
\begin{lemma}[Informal; Sec.~2.3 and Sec.~3.3 of~\citet{liu2022flow}]
    \label{lemma:stochastic-process-flow-matching}
    Let the stochastic process $X(\tau)$ that depends on the noise levels connect the initial random variable $X^0$ and the target random variable $X^1$, i.e., $X(0) = X^0$ and $X(1) = X^1$. Let $\gX$ collect all random variables defining the stochastic process, i.e., $X(\tau)$ is a deterministic function of noise level $\tau$ given $\gX$. Given a noisy sample $x^\tau$ at noise level $\tau$, the velocity field defined as
    $
        u(x^\tau \mid \tau) \triangleq \E_{\gX} \left[ \frac{\de}{\de \tau} X(\tau) \, \middle| \, X(\tau) = x^\tau \right]
    $
    generates~\citep{lipman2022flow} the probability density $p(x^\tau \mid \tau)$ of the stochastic process. In addition, we can use a parametric velocity $v(x^\tau \mid \tau)$ to fit $u(x^\tau \mid \tau)$ by minimizing
    \begin{footalign}
        \E_{\gX, \tau \sim \textsc{Unif}([0, 1])} \left[ \norm*{ v(x^\tau \mid \tau) - \frac{\de}{\de \tau} X(\tau) }^2 \right].
        \label{eq:stoch-proc-flow-matching-loss}
    \end{footalign}
\end{lemma}
Importantly, we can interpret the flow path $\phi(x^0 \mid \tau)$ as a stochastic process, where $u(x^\tau \mid \tau)$ is the velocity and $p(x^\tau \mid \tau)$ is the probability path. We can recover the standard conditional flow matching loss (Eq.~\ref{eq:standard-flow-matching-loss}) by setting $\gX=(X^0, X^1)$ and $X(\tau) = (1 - \tau) X^0 + \tau X^1$. We will use this lemma to construct the flow matching objectives for fitting return distributions in Sec.~\ref{subsec:learning-flow-vels}.

\section{Bellman consistency at all noise levels}
\label{sec:bellman-consistency}

Our goal is to extract richer learning signals from each transition in the data. To achieve this goal, we incorporate Bellman consistency throughout the return flow paths by introducing a novel flow Bellman operator that depends on the noise level (Sec.~\ref{subsec:flow-bellman-eq}). 
We first derive this operator below, and show that it is consistent with the standard Bellman operator at the flow endpoints and has a unique fixed point at all noise levels.
We then use this flow Bellman operator to derive practical flow-matching losses in Sec.~\ref{subsec:learning-flow-vels}.

\subsection{Flow Bellman Equation}
\label{subsec:flow-bellman-eq}

To extend the Bellman equation to the entire flow path, we first define the noisy return random variable at noise level $\tau$ as $Z^{\pi}(s, a, \tau) \in \gZ$. At $\tau=1$, $Z^{\pi}(s, a, 1) \overset{d}{=} Z^{\pi}(s, a)$. Then we introduce a novel \textit{flow Bellman operator} $\gT^{\pi}_\tau: \gZ \to \gZ$, which also depends on the noise level $\tau$ and satisfies the \textbf{\emph{flow Bellman equation}}:
\begin{align}
    Z^{\pi}(s, a, \tau) &\overset{d}{=} \gT^{\pi}_{\tau} Z^{\pi}(s, a, \tau) \overset{d}{=} \tau r(s, a) + \gamma Z^{\pi}(S', A', \tau) + (1 - \tau) \sqrt{1 - \gamma^2} N.
    \label{eq:flow-bellman-eq}
\end{align}
Compared to the distributional Bellman equation in Eq.~\ref{eq:dist-bellman-eq}, this incorporates a standard Gaussian $N \sim \gN(0, 1)$ that is \emph{independent} of the initial noise distribution. This Gaussian correction\footnote{Using other distributions to sample the initial noise, e.g., $\textsc{Unif}([0, 1])$, results in different correction terms.} accounts for the variance in the discounted return at the next time step. To understand this intuitively, we can examine the equation at the two ends of the flow path. At $\tau = 0$, the flow Bellman equation becomes $Z^{\pi}(s, a, 0) \overset{d}{=} \gamma Z^{\pi}(S', A', 0) + \sqrt{1 - \gamma^2} N$. Note that both $Z^{\pi}(s, a, 0)$ and $Z^{\pi}(S', A', 0)$ are standard Gaussian because they are initial flow noises, so the Gaussian correction $N$ is essential for the equation to hold. Omitting the $\sqrt{1-\gamma^2}N$ term, as done in prior work~\citep{dong2025value}, results in an incorrect equation. At the other end of the flow path at $\tau=1$, we recover the standard distributional Bellman equation in Eq. \ref{eq:dist-bellman-eq}.

\vspace{-0.5em}
\paragraph{Unique fixed point at all noise levels.}
We next show that our flow Bellman operator admits a unique fixed point at every noise level ($\tau \in [0, 1]$). This allows us to impose Bellman consistency along the entire flow path, providing richer supervision signals. We start by showing that $\gT^{\pi}_\tau$ is a $\gamma$-contraction.
\begin{remark}[\textbf{Contraction}]
\label{thm:contraction}
For every noise level $\tau$ and a $p \in [1, \infty)$, $\gT^{\pi}_\tau$ is a $\gamma$-contraction on two return random variables, $Z_1$ and $Z_2$, in $\gZ$:
\begin{align}
    W^{\text{sup}}_p ( \gT^{\pi}_\tau Z_1, \gT^{\pi}_\tau Z_2 ) \leq \gamma W_p^{\text{sup}}( Z_1, Z_2 )
\end{align}
where $W_p^{\text{sup}}$ is the supremum of the $p$-Wasserstein distance over state-action pairs.
\end{remark} 
We provide a formal statement and a proof in Appendix~\ref{appendix:flow-bellman-op}. 
Since $\gT^{\pi}_\tau$ is a $\gamma$-contraction, it follows from the contraction mapping theorem~\citep{banach1922operations} that our flow Bellman operator admits the unique fixed point $Z^{\pi}(s, a, \tau)$. In other words, repeatedly applying $\gT^{\pi}_\tau$ to an arbitrary member of $\gZ$ will converge to this fixed point. 
With such a guarantee, we have shown that our proposed flow Bellman operator is well-defined and can provide a consistent bootstrapping target for value-learning. Next, we derive practical flow matching objectives using this operator.

\vspace{-0.5em}
\subsection{Learning return distributions with the flow Bellman equations}
\label{subsec:learning-flow-vels}
\vspace{-0.5em}

In Sec. \ref{subsec:flow-bellman-eq}, we introduced a flow Bellman operator that can enforce Bellman consistency at every noise level of flow matching. However, it is not straightforward how one can use this operator to model $V$- and $Q$-return distributions in practice. In this section, we derive the practical flow-matching losses by first defining the $V$- and $Q$-return random variables and then invoking Lemma~\ref{lemma:stochastic-process-flow-matching}.

We denote the $V$-return random variable as $Z^{\pi}_V(s, \tau)$ and the $Q$-return random variable as $Z^{\pi}_Q(s, a, \tau)$. We can decompose the flow Bellman equation (Eq.~\ref{eq:flow-bellman-eq}) into two parts:
\begin{equation}
    Z_V^{\pi}(s, \tau) \overset{d}{=} Z_Q^{\pi}(s, A, \tau),  \quad 
    Z_Q^{\pi}(s,a,\tau) \overset{d}{=} \tau r(s, a) + \gamma Z_V^{\pi}(S', \tau) + (1 - \tau) \sqrt{1 - \gamma^2} N,
    \label{eq:flow-bellman-eqs}
\end{equation}
where $A$ is the action random variable at the current time step. 
The first equation relates $Z^\pi_V(s, \tau)$ to $Z^\pi_Q(s, A, \tau)$ via actions sampled from the policy, and the second equation corresponds to the temporal-difference backup as in Eq.~\ref{eq:flow-bellman-eq}. Our goal is to model these two random variables using flow matching. Specifically, we want to use flow paths $\phi^{\pi}_V$ and $\phi^{\pi}_Q$ to transport independent Gaussian random variables $N_V, N_Q$ into these laws:
\begin{align}
    \phi^{\pi}_V(N_V \mid s, \tau) &\overset{d}{=} \phi^{\pi}_Q(N_Q \mid s, A, \tau), \label{eq:value-flow} \\
    \phi^{\pi}_Q(N_Q \mid s, a, \tau) &\overset{d}{=} \tau r(s, a) + \gamma \phi^{\pi}_V(N_V \mid S', \tau) + (1 - \tau) \sqrt{1 - \gamma^2} N,
    \label{eq:critic-flow}
\end{align}
where $\phi^{\pi}_V$ and $\phi^{\pi}_Q$ solve their corresponding ODEs (Eq.~\ref{eq:flow-ode}) with velocities $u^{\pi}_V$ and $u^{\pi}_Q$, respectively.

\vspace{-0.5em}
\paragraph{Learning the $V$-flow $\phi^{\pi}_V$.} 
The identity in Eq.~\ref{eq:value-flow} suggests that $\phi^{\pi}_V$ is a marginal distribution over the Gaussian random variable $N_Q$ and the action random variable $A$. We can combine these random variables into the set $\gX_V = \{N_Q, A \}$ and use it to determine the stochastic process $X^{\pi}_V(s, \tau) \triangleq \phi^{\pi}_Q(N_Q \mid s, A, \tau)$ that corresponds to the $V$-flow $\phi^{\pi}_V$. Taking noise-level differentiation, we have $\frac{\de}{\de \tau} X^{\pi}_V(s, \tau) = u^{\pi}_Q( \phi^{\pi}_Q(N_Q \mid s, A, \tau) \mid s, A, \tau)$, where we plug in the ODE between $\phi^{\pi}_Q$ and $u^{\pi}_Q$ as in Eq.~\ref{eq:flow-ode}. By Lemma~\ref{lemma:stochastic-process-flow-matching}, defining the stochastic process reveals the velocity of the $V$-flow, and we can take expectations over $\gX_V$ and use a parametric $V$-velocity $v_V(z^{\tau} \mid s, \tau)$ to fit $u^{\pi}_V(z^\tau \mid s, \tau)$ by minimizing
\begin{align}
    \E_{ \substack{a \sim \pi(a \mid s), \: \tau \sim \textsc{Unif}([0, 1]), \\ z^0 \sim \gN(0, 1), \: z^\tau = \phi^{\pi}_Q(z^0 \mid s, a, \tau) } } \left[ (v_V(z^\tau \mid s, \tau) - u^{\pi}_Q(z^\tau \mid s, a, \tau) )^2 \right].
    \label{eq:value-ret-fm-prelim-loss}
\end{align}

\vspace{-0.5em}
\paragraph{Learning the $Q$-flow $\phi^{\pi}_Q$.}
Similarly, the identity in Eq.~\ref{eq:critic-flow} suggests that $\phi^{\pi}_Q$ is a marginal distribution over Gaussian random variables $N_V, N$ and the next state $S'$. Thus, we can set $\gX_Q = \{ N_V, N, S' \}$ and define the stochastic process $X^{\pi}_Q(s, a, \tau)$ that corresponds to the $Q$-flow with noise-level differentiation as
\begin{align*}
    X^{\pi}_Q(s, a, \tau) &\triangleq \tau r(s, a) + \gamma \phi^{\pi}_V(N_V \mid S', \tau) + (1 - \tau) \sqrt{1 - \gamma^2} N, \\
    \frac{\de}{\de \tau} X^{\pi}_Q(s, a, \tau) &= r(s, a) + \gamma u^{\pi}_V( \phi^{\pi}_V(N_V \mid S', \tau) \mid S', \tau) - \sqrt{1 - \gamma^2} N.
\end{align*}
Taking expectations over $\gX_Q$, we can also use a parametric $Q$-velocity $v_Q(z^\tau \mid s, a, \tau)$ to fit $u^{\pi}_Q(z^{\tau} \mid s, a, \tau)$ by minimizing
\begin{align}
    &\E_{ \substack{s' \sim p(s' \mid s, a), \: \tau \sim \textsc{Unif}([0, 1]), \\
    z^0, \: \epsilon \: \overset{\text{iid}}{\sim} \: \gN(0, 1), \: z^\tau = \phi^{\pi}_V(z^0 \mid s', \tau) }} \left[ \left( v_Q \left( \tau r + \gamma z^\tau + (1 - \tau) \sqrt{1 - \gamma^2} \epsilon \, \middle| \, s, a, \tau \right) \right. \right. \nonumber \\
    &\hspace{12.5em} \left. \left. - \left( r + \gamma u^{\pi}_V(z^\tau \mid s', \tau) - \sqrt{1 - \gamma^2} \epsilon \right) \right)^2 \right].
    \label{eq:critic-ret-fm-prelim-loss}
\end{align}

\vspace{-0.5em}
\paragraph{Practical objectives.}
\label{subsec:regression-objs}
Note that both losses we derived in Eq.~\ref{eq:value-ret-fm-prelim-loss} and Eq.~\ref{eq:critic-ret-fm-prelim-loss} require access to ground-truth velocities $u^{\pi}_V$ and $u^{\pi}_Q$.
Since ground-truth velocities are unknown, we learn parametric velocities $v_V, v_Q$ using each other as targets. This procedure resembles Fitted Q-Evaluation~\citep{riedmiller2005neural, le2019batch} for learning the scalar $V$ and $Q$ functions. 
In practice, we additionally scale the reward function by $(1-\gamma)$ to normalize the scale of the discounted future return in an infinite-horizon MDP, helping to boost performance (see Sec.~\ref{subsec:design-decisions} for empirical studies).
This yields losses:
\begin{greenbox}[title={Practical objectives (\methodexplain{})}]
\begin{footalign}
    \gL(v_V) &= \E_{ \substack{ a \sim \pi(a \mid s), \tau \sim \textsc{Unif}([0, 1]), \\ z^\tau \sim p_Q(z^\tau \mid s, a, \tau) } } \left[ \left( v_V(z^\tau \mid s, \tau) - v_Q(z^\tau \mid s, a, \tau) \right)^2 \right], \label{eq:value-ret-fm-loss} \\
    \gL(v_Q) &= \E_{ \substack{s' \sim p(s' \mid s, a), \: \tau \sim \textsc{Unif}([0, 1]), \\
    \epsilon \sim \gN(0, 1), \: z^\tau \sim p_V(z^{\tau} \mid s', \tau ) }} \left[ \left( v_Q \left( \tau(1 - \gamma) r + \gamma z^\tau + (1 - \tau) \sqrt{1 - \gamma^2} \epsilon \, \middle| \, s, a, \tau \right) \right. \right. \nonumber \\
    &\hspace{12.5em} \left. \left. - \left( (1 - \gamma) r + \gamma v_V(z^\tau \mid s', \tau) - \sqrt{1 - \gamma^2} \epsilon \right) \right)^2 \right],
    \label{eq:critic-ret-fm-loss}
\end{footalign}
where $z^\tau \sim p_Q(z^\tau \mid s, a, \tau)$ and $z^\tau \sim p_V(z^\tau \mid s', \tau)$ denote the sampling of the noisy returns.
\end{greenbox}
We alternate between minimizing $\gL(v_V)$ and $\gL(v_Q)$ iteratively, holding the other component fixed during each update. Minimizing these losses performs policy evaluation to find the return flows under policy $\pi$ at each noise level, and we call these recipes \textbf{\emph{\methodexplain{}}}. In the next section, we will discuss two strategies to find the optimal policy by explicit or implicit policy improvement based on the \methodexplain{} losses.

\section{\methodexplain{} algorithms}
\label{sec:value-forcing-algorithms}

We consider two ways to turn the richer distributional learning signals of \methodexplain{} into RL algorithms. 
First, we can evaluate an explicit policy by learning its return distributions and then performing policy improvement. We refer to this approach as the \textit{explicit} method, and introduce a practical algorithm named \methodname{}. Second, motivated by prior work on implicit value learning~\citep{kostrikov2022offline, sikchi2024dual}, we learn return distributions through an expectile-weighted flow matching objective without an explicit actor. 
We refer to this type of approach as the \textit{implicit} method, and introduce an implicit variant of our method called \methodnamei{} (Implicit VF).

\label{subsec:value-forcing}
\textbf{Explicit method.}
As in the standard actor-critic framework, the objective is to learn a policy that maximizes the $Q$ estimate. In the context of return flows, $Q$-values are estimated as: $Q(s, a) \triangleq \E_{z^0 \sim \gN(0, 1)} \left[ \int_0^1 v_Q(z^\tau \mid s, a, \tau) d\tau \right]$,
where we use $M$ noise samples to compute this quantity. We instantiate $Q$ maximization as rejection sampling~\citep{chen2023offline} on $K$ actions proposed by a flow behavioral cloning (BC) policy $\pi_\beta(a \mid s)$:
\begin{align}
     \pi(a \mid s) \overset{d}{=} \argmax_{a_1, \cdots, a_K: \: a_i \sim \pi_\beta(a \mid s)} Q(s, a_i).
     \label{eq:flow-rej-sampling}
\end{align}
We learn the BC policy using the standard flow-matching loss (Eq.~\ref{eq:standard-flow-matching-loss}) and denote the loss as $\gL(\pi_\beta)$. The policy $\pi$ is then used to sample actions for the objective in Eq.~\ref{eq:value-ret-fm-loss}, thereby coupling policy evaluation with action selection under the updated $Q$-flow.

\textbf{Implicit method.}
Another instantiation of our method, \methodnamei{}, avoids explicit actors and only queries actions within the dataset during training. 
Specifically, we replace the actions sampled from an explicit policy in Eq.~\ref{eq:value-ret-fm-loss} with dataset actions and then add an expectile weight into the regression objectives. 
Following prior work (e.g., Sec.~10.3.1 of~\citet{fahrmeir2022regression}), we define the expectile weight with expectile $\mu \in [0.5, 1)$ as
\begin{align}
    w^\mu(s, a) \triangleq \mu \mathbbm{1} \left( Q(s, a) \geq V(s) \right) + (1 - \mu) \mathbbm{1}( Q(s, a) < V(s)),
    \label{eq:expectile-weight}
\end{align}
where $V(s) \triangleq \E_{z^0 \sim \gN(0, 1)} \left[ \int_0^1 v_V(z^\tau \mid s, \tau) d\tau \right]$ is the $V$-estimate.
The resulting expectile-weighted regression objective is 
\begin{align}
    \gL^\mu(v_V) = \E_{ \substack{ a \sim D, \: \tau \sim \textsc{Unif}([0, 1]), \\ z^\tau \sim p_Q(z^\tau \mid s, a, \tau) } } \left[ w^\mu(s, a) (v_V(z^{\tau} \mid s, \tau) - v_Q(z^\tau \mid s, a, \tau) )^2 \right].
    \label{eq:value-ret-imp-vel-loss}
\end{align}
This loss extends the scalar expectile regression loss~\citep{kostrikov2022offline} to return flows. In Appendix~\ref{appendix:ivf-proofs}, we show that \emph{(1)} minimizing the expectile-weighted loss induces a $V$-estimate that is the expectile of the $Q$-estimate, and \emph{(2)} when the expectile $\mu \rightarrow 1$, the value estimate converges to the optimal value within the dataset support.

To summarize, \methodname{} and \methodnamei{} consist of a $V$-velocity $v_V(z^\tau \mid s,\tau)$, a $Q$-velocity $v_Q(z^\tau \mid s,a,\tau)$, and a flow BC policy
$\pi_\beta(a \mid s)$. We use neural networks to parameterize
the $V$-velocity $v_{V}^{\theta}$, $Q$-velocity
$v_{Q}^{\psi}$, and flow BC velocity $v_{\beta}^{\xi}$.
The scalar estimates $V(s)$ and $Q(s,a)$ are obtained by averaging
$M$ return samples generated by their return flows. At evaluation time, both algorithms select actions using rejection sampling over $K$ candidates from $\pi_\beta$. \methodname{} also uses this action selection procedure to learn the $V$-velocity, while \methodnamei{} uses actions in the dataset to perform expectile-weighted regression. Alg.~\ref{alg:value-forcing} summarizes the training procedures of {\color{mygreen} \methodname{}} and {\color{mygreendark} \methodnamei{}} and our open-source implementation is available at \url{https://github.com/chongyi-zheng/value-forcing}.

\begin{algorithm}[t]
\caption{{\color{mygreen} Value Forcing (\methodname{})} and
{\color{mygreendark} Implicit Value Forcing (\methodnamei{})}}
\label{alg:value-forcing}
\algrenewcommand{\alglinenumber}[1]{\textcolor{black}{\footnotesize #1:}}
\begin{algorithmic}[1]
    \Require Dataset $D$, discount $\gamma$, {\color{mygreendark} expectile $\mu$},
    number of action candidates $K$, number of MC return samples $M$.

    \State Initialize velocities $v_{V}^{\theta}, v_{Q}^\psi, v_{\beta}^\xi$ and initialize target velocity $\bar\psi \gets \psi$.

    \For{each training iteration}
        \State $(s, a, r, s') \sim D$, $\tau \sim \text{Unif}([0, 1])$.

        {\color{mygreen}
            \State $a_\pi \sim$ rejection sampling
            (Eq.~\ref{eq:flow-rej-sampling}),
            $z^\tau \sim p_Q^{\bar{\psi}}(z^\tau \mid s,a_\pi,\tau)$.
            \State $\Delta\theta \leftarrow
            \nabla_\theta \gL(v_V^\theta)$
            (Eq.~\ref{eq:value-ret-fm-loss}).
            \TwoLineComment[mygreen]{Learn $V$-velocity for \methodname{}}
        }
        
        {\color{mygreendark}
            \State Compute $w^\mu(s,a)$ using Eq.~\ref{eq:expectile-weight},
            $z^\tau \sim p_Q^{\bar{\psi}}(z^\tau \mid s,a,\tau)$.
            \State $\Delta\theta \leftarrow
            \nabla_\theta \gL^\mu(v_V^\theta)$
            (Eq.~\ref{eq:value-ret-imp-vel-loss}).
            \TwoLineComment[mygreendark]{Learn $V$-velocity for \methodnamei{}}
        }

        \State $z^\tau \sim p_{V}^{\theta}(z^\tau \mid s',\tau)$, $\epsilon \sim \gN(0,1)$.
        \State $\Delta \psi \leftarrow \nabla_{\psi} \gL(v_{Q}^{\psi})$ (Eq.~\ref{eq:critic-ret-fm-loss}).
        \TwoLineComment[mygray]{Learn $Q$-velocity}

        \State $\Delta \xi \leftarrow \nabla_{\xi} \gL(v_\beta^{\xi})$ (Eq.~\ref{eq:standard-flow-matching-loss}). \Comment{Learn flow BC policy}

        \State $\bar\psi \leftarrow$ the exponential moving average of $\psi$. \Comment{Update the target parameters}
    \EndFor
\end{algorithmic}
\end{algorithm}

\section{Experiments}
\label{sec:experiments}

We conduct experiments to study how well \methodexplain{} can learn the return distribution on a range of challenging long-horizon tasks. Concretely, we ask the research questions: \textbf{(1)} Does \methodexplain{} improve return estimation accuracy? \textbf{(2)} How do \methodname{} and \methodnamei{} compare to prior offline RL methods across different dataset sizes? \textbf{(3)} What are the important design decisions in \methodname{} and \methodnamei{}?

\textbf{Experimental setup.} We evaluate our methods on $20$ tasks from OGBench~\citep{park2024ogbench} and $6$ tasks from OCBench~\citep{park2026behavioral} in the offline RL setting. OGBench includes $4$ domains with $5$ tasks each: \texttt{cube-triple}, \texttt{cube-quadruple}, \texttt{puzzle-4x4}, and \texttt{humanoidmaze-giant}. OCBench includes $1$ task from \texttt{block-lite-single}, $2$ from \texttt{block-lite-double}, and $3$ from \texttt{chamber-lite-easy}. We adopt action chunking~\citep{li2025reinforcement}, where the policy predicts and executes a sequence of actions before predicting the next sequence. The resulting higher-dimensional action space makes value learning more challenging. Following~\citet{zhou2026test}, we use the OGBench datasets with $100$M transitions as the full offline datasets and experiment with subsets of them, ranging from $10\%$ to $100\%$, to study the effectiveness of different value learning methods on decreasing dataset sizes. For OCBench, we compare our methods with selected baselines using $25\%$ and $100\%$ of the available data. We report the mean and standard deviation across $4$ random seeds for all experiments. See Appendix~\ref{appendix:data-and-eval-proto} for details on dataset construction and evaluation protocols.

\textbf{Comparisons.} We compare \methodname{} and \methodnamei{} with prior value learning methods, grouped into explicit or implicit methods. For \emph{explicit} methods, we compare \methodname{} against prior methods that explicitly learn a policy using flow-based distributional critics \textbf{FlowCritic}~\citep{zhong2025flowcritic}, \textbf{Value Flows}~\citep{dong2025value}, \textbf{PCBF}~\citep{xu2026path}, categorical distributional critic \textbf{C51}~\citep{bellemare2017distributional}, quantile distributional critic \textbf{IQN}~\citep{dabney2018implicit}, flow-based non-distributional critic \textbf{Floq}~\citep{agrawalla2026floq}, and scalar critic \textbf{QC}. For \textit{implicit methods}, we compare \methodnamei{} against prior methods that perform implicit value learning. Since there are no prior implicit flow-based distributional methods, we compare against categorical distributional critics \textbf{DIVL}~\citep{wang2026learning}, our novel adaptations of C51 and IQN termed \textbf{Implicit C51} and \textbf{Implicit IQN}, and scalar critic \textbf{Implicit QC}. For OCBench experiments, we compare \methodname{} with representative baselines Value Flows, C51, and QC, and compare \methodnamei{} with Implicit C51 and Implicit QC. See Appendix~\ref{appendix:prior-methods} for details of each prior method.

\subsection{\methodexplain{} improves return estimation accuracy}
\label{subsec:ret-hist}

\begin{figure}[t]
    \vspace{-3em}
    \centering
    \includegraphics[width=0.8\linewidth]{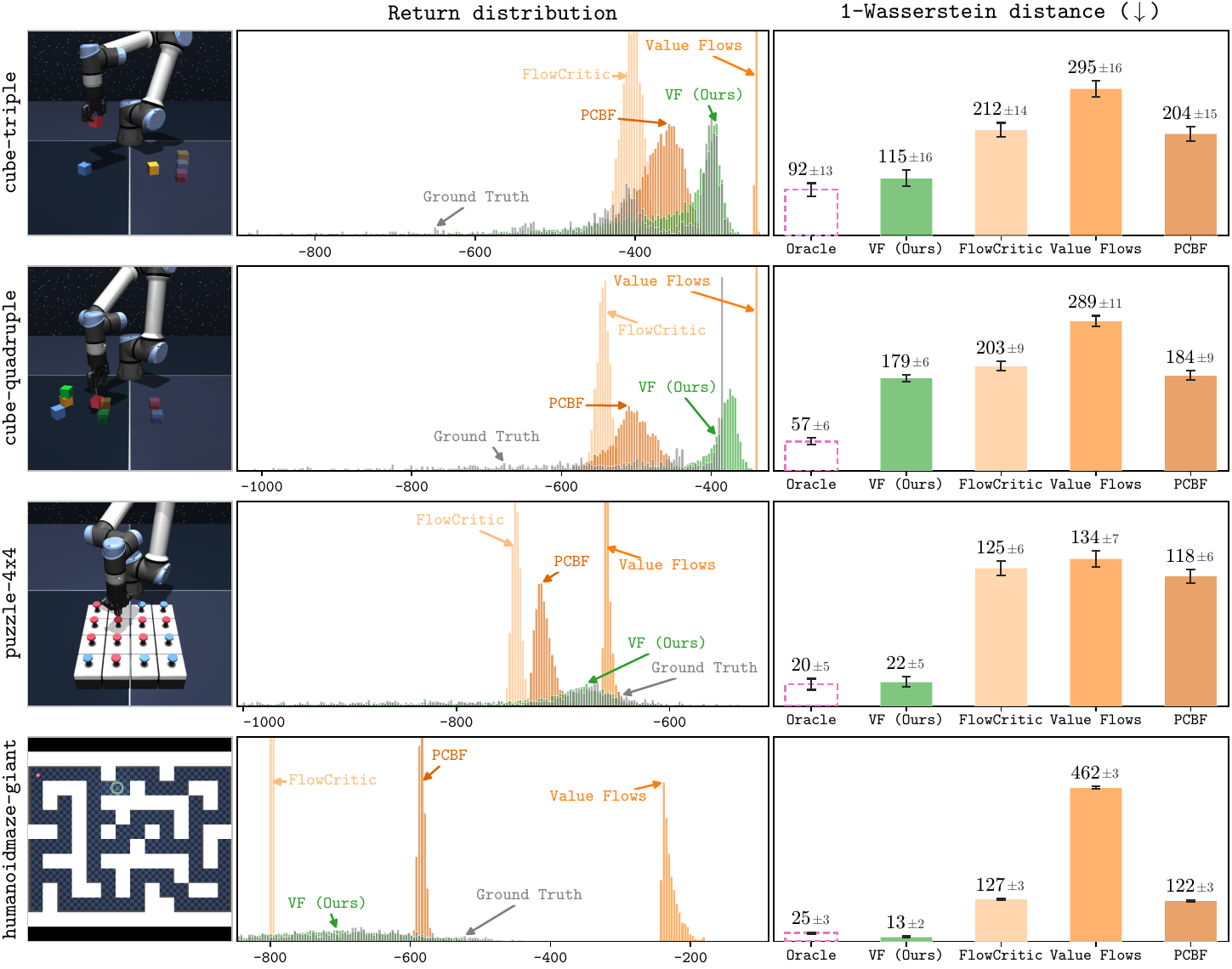}
    \caption{\footnotesize\textbf{\methodexplain{} learns more accurate return distributions than prior flow-based return estimators.} 
    \figleft{} We collect datasets to conduct policy evaluation experiments at a \emph{single} state from each of the four OGBench domains. \figcenter{} \methodexplain{} captures the ground-truth return distributions more accurately, while prior methods are biased and collapse their predictions into narrow peaks.
    \figright{} \methodexplain{} achieves lower mean $1$-Wasserstein distances than all three prior methods in all tasks. 
    }
    \vspace{-2em}
    \label{fig:ret-hist}
\end{figure}

We investigate whether \methodexplain{} improves return distribution prediction by conducting
\emph{policy evaluation} experiments comparing \methodname{} with Value Flows, FlowCritic, and PCBF at a single state randomly sampled from each of the four OGBench domains.
Our goal is to compare the return estimation accuracy starting from a single state in each domain and then following a fixed policy.
Specifically, from each state, we execute a fixed starting action chunk and then roll out a data-collection policy to collect $1000$ training trajectories and $1000$ evaluation trajectories separately. We train each method on the training trajectories, using the recorded actions in the bootstrap targets, effectively performing SARSA~\citep{sutton1998reinforcement}. For evaluation, we use discounted returns from the evaluation trajectories as the ground truth. Thus, all methods evaluate the same data-collection policy. Following~\citet{dong2025value}, we visualize the predicted return distributions and measure their $1$-Wasserstein distances to the ground-truth returns. As a reference, an \texttt{Oracle} return estimator fits the realized returns of the same trajectories using the standard flow-matching loss (Eq.~\ref{eq:standard-flow-matching-loss}). This provides a reference for errors other than bootstrapping: function approximation, optimization, finite samples, etc. See Appendix~\ref{appendix:ret-hist} for the full experimental details.

Fig.~\ref{fig:ret-hist} shows that \methodname{} achieves lower mean $1$-Wasserstein distances than previous flow-based critics on all four tasks. Observe that \methodname{} captures more of the ground-truth spread, while Value Flows and FlowCritic concentrate their predictions into narrow peaks. PCBF produces broader predictions on the \texttt{cube} tasks, but still misses much of the ground-truth spread. Relative to PCBF, \methodname{} reduces the mean $1$-Wasserstein distance by $81\%$ on \texttt{puzzle-4x4} and $89\%$ on \texttt{humanoidmaze-giant}. On \texttt{cube-triple} and \texttt{puzzle-4x4}, \methodname{}'s $1$-Wasserstein distances are also close to the \texttt{Oracle}. 
In particular, \methodname{} corrects Value Flows' inconsistent supervision at $\tau = 0$, offering an explanation for the improvement over its narrow, shifted predictions. 
In Appendix Fig.~\ref{fig:ret-flows}, we visualize the return flow paths at different noise levels, showing that \methodname{} spreads its return paths at early noise levels (around $\tau = 0.25$), while baselines collapse (around $\tau = 0.9$). The gains over FlowCritic and PCBF further suggest the benefit of enforcing Bellman consistency of noisy return distributions at every noise level.

\subsection{\methodname{} and \methodnamei{} outperform prior offline RL methods on smaller datasets}
\label{subsec:offline-exps}

\begin{figure}[t]
    \vspace{-3em}
    \centering
    \includegraphics[width=\linewidth]{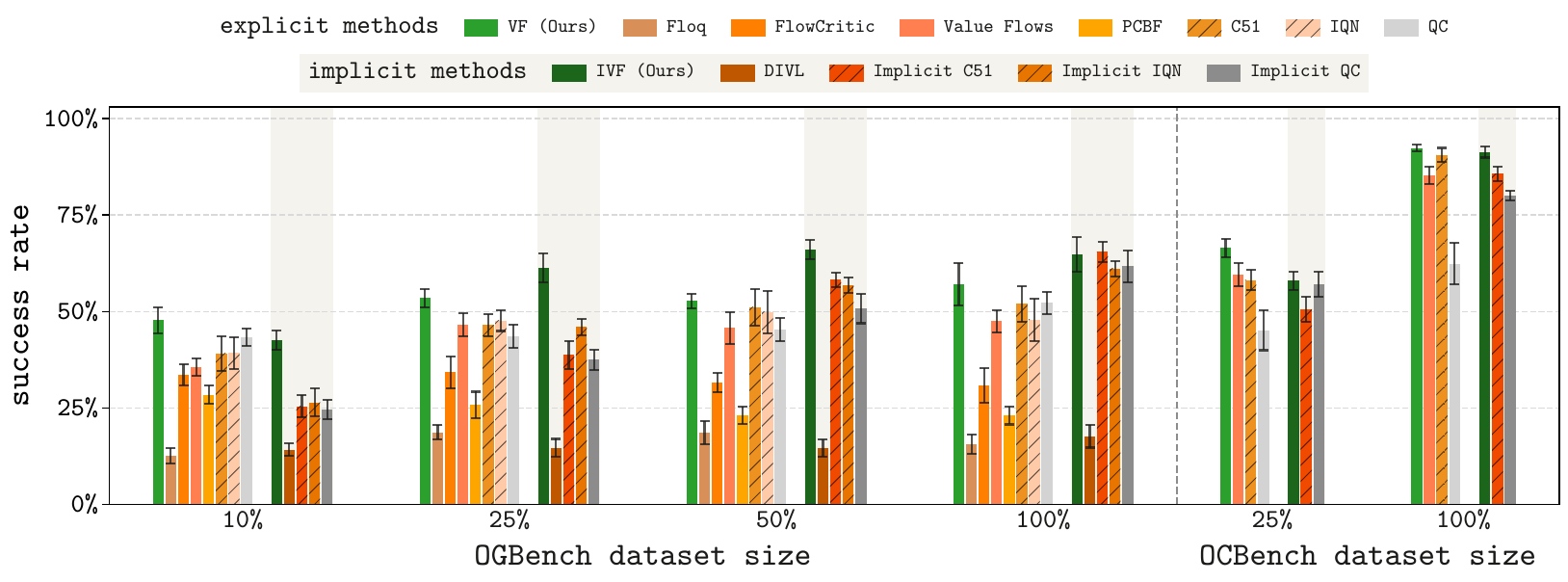}
    \vspace{-2em}
    \caption{\footnotesize
    \textbf{\methodexplain{} improves offline RL performance across dataset sizes.} Success rates are averaged over $20$ OGBench tasks \figleft{} and $6$ OCBench tasks \figright{} at each dataset size. On OGBench, \methodname{} and \methodnamei{} outperform their explicit and implicit baselines at $10\%$ and $25\%$ of the data, and remain competitive with more data. On OCBench, \methodname{} outperforms selected explicit methods on $25\%$ of the dataset, while \methodnamei{} leads among selected implicit methods on the full dataset. Hatched bars (\distRLlegend{}) indicate distributional RL methods without flow matching. See Appendix~\ref{appendix:complete-results} for complete results.
    }
    \label{fig:offline-results-agg}
    \vspace{-2em}
\end{figure}

We next compare \methodname{} and \methodnamei{} against prior value-learning algorithms under varying dataset sizes to evaluate different methods as data get increasingly limited. On OGBench, we train each method using the full dataset as well as subsets containing $50\%$, $25\%$, and $10\%$ of the available transitions. We construct these subsets by sampling trajectories using a softmax over their average rewards (Appendix~\ref{appendix:data-and-eval-proto}). We additionally evaluate the methods on OCBench using $25\%$ and $100\%$ of the available data. Fig.~\ref{fig:offline-results-agg} summarizes the success rates on both benchmarks, grouping the methods into explicit and implicit methods.

On OGBench, \methodname{} achieves the highest mean success rates with $10\%$ and $25\%$ of the data and performs similarly to the strongest baseline methods at larger dataset sizes. Most prior flow-based distributional RL methods significantly underperform \methodname{}, and even the scalar critic QC. This difference is especially prominent with $10\%$ of the data, highlighting the importance of Bellman consistency at all noise levels; not doing so makes the flow-based distributional baselines perform even worse than the scalar critic.
In contrast, \methodname{} barely performs worse as the dataset size decreases. Among implicit methods, \methodnamei{} performs comparably to the strongest baselines on the full dataset and achieves higher mean success rates than all prior methods at every reduced dataset size. On OCBench, \methodname{} achieves the highest mean success rates among the evaluated explicit methods with both $25\%$ and $100\%$ of the data, while \methodnamei{} leads among implicit methods on the full dataset and performs comparably to the baselines with $25\%$ of the data. These results suggest that aligning return distributions at all noise levels is incredibly important for training flow-based distributional critics, and is especially beneficial when offline data is limited.

\subsection{Key design decisions in \methodexplain{}}
\label{subsec:design-decisions}

\begin{figure}[t]
    \centering
    \vspace{-1.0cm}
    \begin{minipage}[t]{0.49\textwidth}
        \centering
        \includegraphics[width=\textwidth]{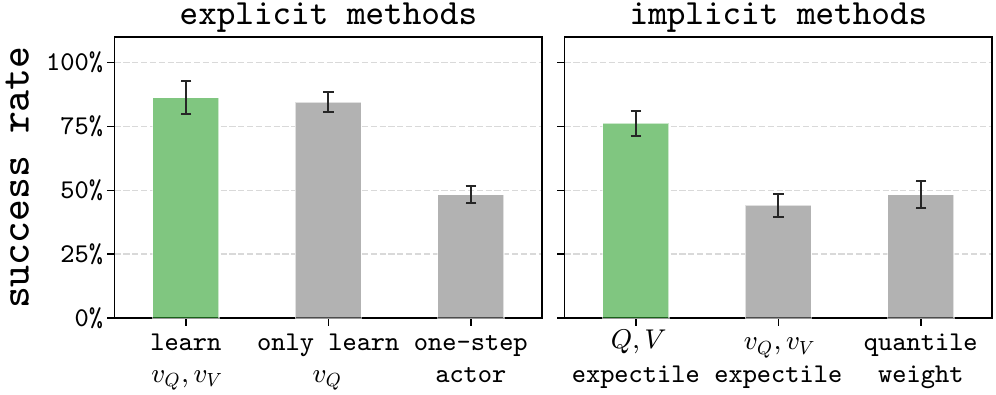}
        \vspace{-2em}
        \caption{
        Comparisons of alternative \methodexplain{} instantiations for explicit methods (Left) and implicit methods (Right).
        }
        \label{fig:value-policy-ablations}
    \end{minipage}
    \hfill
    \begin{minipage}[t]{0.49\textwidth}
        \centering
        \includegraphics[width=\textwidth]{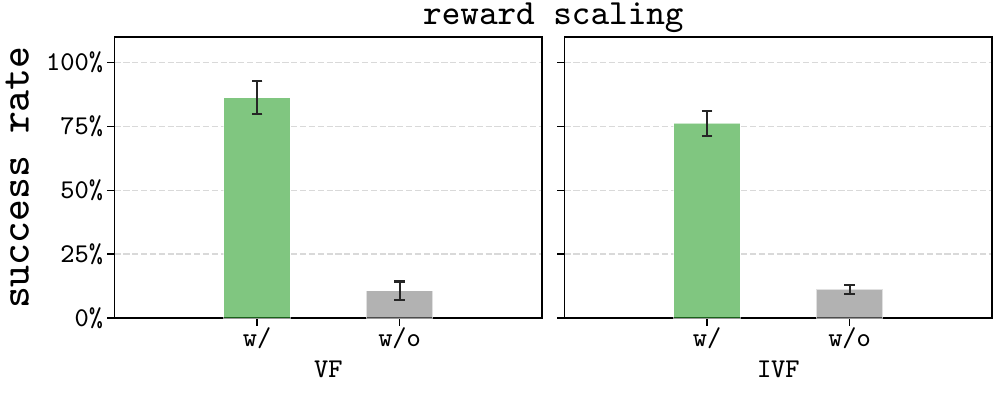}
        \vspace{-2em}
        \caption{\footnotesize \textbf{Reward scaling improves \methodname{} and \methodnamei{}.} Both \methodname{} and \methodnamei{} achieve substantially higher success rates with reward scaling (\texttt{w/}) than without it (\texttt{w/o}).}
        \label{fig:reward-scaling-ablations}
    \end{minipage}
    \vspace{-2em}
\end{figure}

Finally, we study key design decisions for \methodname{} and \methodnamei{} by ablating them on tasks $2$ and $4$ from \texttt{cube-triple} and \texttt{puzzle-4x4}. These ablations highlight the importance of policy extraction, expectile weighting, and reward scaling, while showing flexibility in the choice of bootstrap update for \methodname{}. Appendix~\ref{appendix:alternative-algos} provides experimental details and descriptions of each ablation.

For \methodname{}, we ablate two alternative designs: removing the separate $V$-velocity in Eq.~\ref{eq:critic-ret-fm-loss} and instead using $Q$-velocity to bootstrap itself (\texttt{$v_Q$ only}), and replacing rejection sampling with a learned one-step actor that maximizes the Q estimate~\citep{park2025flow} (\texttt{one-step actor}). Fig.~\ref{fig:value-policy-ablations} \figleft{} shows that only learning $v_Q$ yields similar success rates to \methodname{}, whereas using a one-step actor reduces success rates by $44\%$. These results suggest that \methodname{} learning a separate V-velocity is optional, while the choice of policy extraction has a larger effect on \methodname{}.

For \methodnamei{}, we examine alternatives to the expectile weighting in the $V$-velocity loss (Eq.~\ref{eq:value-ret-imp-vel-loss}). We compare weighting based on scalar advantages (\texttt{$Q, V$ expectile}) against applying expectile regression directly to velocity residuals (\texttt{$v_Q, v_V$ expectile}) or weighting based on return quantiles (\texttt{quantile weight}). Fig.~\ref{fig:value-policy-ablations} \figright{} shows that weighting based on scalar advantages achieves much higher success rates than its two counterparts. These results support using scalar advantages to determine the expectile weights when learning return velocities.

Finally, Fig.~\ref{fig:reward-scaling-ablations} examines the effect of reward scaling in Eq.~\ref{eq:critic-ret-fm-loss}. Removing scaling reduces mean success rates from approximately $86\%$ to $11\%$ for \methodname{} and from $76\%$ to $11\%$ for \methodnamei{}, highlighting the importance of reward scaling in the practical implementation of both methods.

\paragraph{Additional experiments.} In Appendix~\ref{appendix:hyperparam-ablations}, we examine the effects of the number of flow integration steps and return samples used to estimate $V$ and $Q$, showing that both \methodname{} and \methodnamei{} maintain similar performance across the tested settings. We also extend the return estimation experiments in Sec.~\ref{subsec:ret-hist} to implicit distributional RL baselines. Appendix Fig.~\ref{fig:implicit-ret-hist} shows that \methodnamei{} better captures the ground-truth return distributions and achieves lower mean $1$-Wasserstein distances than alternative implicit methods.

\section{Conclusion}

In this paper, we propose \methodexplain{}, a new approach to learning distributional value functions with flow matching that enforces Bellman consistency at every noise level. This provides richer supervision during training, enabling more accurate estimation of return distributions and better policy performance, especially in the limited-data regime.

Our results suggest two directions for future work. First, our theoretical analysis of \methodnamei{} assumes exact optimization and convergence to a self-consistent fixed point. Establishing convergence guarantees for practical algorithms with neural network approximation is an open problem. Second, extending \methodexplain{} to online learning and real-world applications, such as robotics, is another promising direction for future work.

\subsection*{Acknowledgement}

We would like to thank Seohong Park for helpful discussions and feedback on early drafts of this paper. This research was partly supported by DARPA ANSR and ONR N00014-25-1-2060. 
This research used the Savio computational cluster resource provided by the Berkeley Research Computing program at UC Berkeley, the Della computational cluster provided by Princeton Research Computing, and the Ionic and the Neuronic computational clusters provided by the Department of Computer Science at Princeton University.

\appendix

\begin{table*}[p]
\centering
\caption{
    \textbf{Comparison of Bellman constraints across flow-based value learning methods.}
    Among the methods compared, \methodname{} and \methodnamei{} model the full return distribution, while enforcing Bellman consistency at every noise level.
    This provides consistent supervision throughout the flow path, improving return estimation (Sec.~\ref{subsec:ret-hist}) and offline RL performance (Sec.~\ref{subsec:offline-exps}), particularly when data are limited.
}
\label{tab:flow-bellman-comparison}
\small
\setlength{\tabcolsep}{2pt}
\renewcommand{\arraystretch}{1.25}

\makebox[\textwidth][c]{%
\scalebox{0.88}{%
\begin{tabular}{@{}lccccccc@{}}
\toprule

\texttt{Method}
&
\methodhead{Floq}{agrawalla2026floq}
&
\methodhead{FlowCritic}{zhong2025flowcritic}
&
\methodhead{FlowIQN}{groom2026quantile}
&
\methodhead{Value Flows}{dong2025value}
&
\methodhead{EVOR}{espinosa2025expressive}
&
\methodhead{PCBF}{xu2026path}
&
\ourmethodhead{\methodname{}/\methodnamei{}}
\\
\midrule

Full return distribution?
&
\xmark
&
\cmark
&
\cmark
&
\cmark
&
\cmark
&
\cmark
&
\cmark
\\

Regression target
&
{\makecell{scalar\\endpoint}}
&
{\makecell{endpoint\\samples}}
&
{\makecell{quantile-coupled\\samples}}
&
{\makecell{return\\velocity}}
&
{\makecell{return\\velocity}}
&
{\makecell{noise-coupled\\velocity}}
&
{\makecell{return\\velocity}}
\\

\makecell[l]{Gaussians at initial\\noise levels?}
&
\xmark
&
\cmark
&
\xmark
&
\xmark$^\dagger$
&
\xmark
&
\cmark
&
\cmark
\\

\makecell[l]{Bellman constraints at\\intermediate noise levels?}
&
\xmark
&
\xmark
&
\xmark
&
\cmark
&
\xmark$^\ddagger$
&
\xmark$^\S$
&
\cmark
\\

\makecell[l]{Bellman constraints at\\final noise levels?}
&
\cmark
&
\cmark
&
\cmark
&
\cmark
&
\cmark
&
\cmark
&
\cmark
\\

\bottomrule
\end{tabular}%
}%
}

\vspace{0.3em}
\begin{minipage}{0.98\textwidth}
\footnotesize
\raggedright
$^\dagger$Value Flows imposes a Bellman equation on intermediate
marginals, but it does not recover Gaussians at the initial noise levels. $^\ddagger$EVOR uses return velocities to construct TD targets directly.
$^\S$PCBF couples two flow paths using shared initial noises rather than enforcing each intermediate marginal distribution to satisfy a Bellman equation.
\end{minipage}
\end{table*}

\section{Proofs and Derivations}
\label{appendix:proofs}

\subsection{Contraction of the flow Bellman operator}
\label{appendix:flow-bellman-op}

\begin{proposition}\label{prop:flow-bellman-contraction} Given a policy $\pi$, a discount factor $\gamma \in [0,1)$, a noise level $\tau \in [0, 1]$, and $p \in [1, \infty)$, for two arbitrary members of the return random variables $Z_1 \in \gZ$ and $Z_2 \in \gZ$, we define the supremum of the $p$-Wasserstein distance over all state-action pairs as
\begin{align*}
    W^{\text{sup}}_p(Z_1, Z_2)
    \triangleq
    \sup_{s, a}
    W_p \left( Z_1(s, a, \tau), Z_2(s, a, \tau) \right),
\end{align*}
where $W_p(Z_1, Z_2)$ is the $p$-Wasserstein distance between the corresponding probability laws of random variables $Z_1$ and $Z_2$~\citep{villani2009optimal}. Then, the flow Bellman operator $\gT^{\pi}_\tau$ satisfies 
\begin{align*}
    W^{\text{sup}}_p \left( \gT_\tau^\pi Z_1, \gT_\tau^\pi Z_2 \right) \leq \gamma W^{\text{sup}}_p( Z_1, Z_2 ).
\end{align*}
Therefore, the flow Bellman operator $\gT^\pi_\tau$ is a $\gamma$-contraction under the metric $W^{\text{sup}}_p$.
\end{proposition}

\begin{proof} Given a state-action pair $(s, a)$, let $S' \sim p(S' \mid s, a)$ be the random variable of the next state and $A' \sim \pi(A' \mid S')$ be the random variable of the next action. Our goal is to find a coupling between random variables $Z_1(S', A', \tau)$ and $Z_2(S', A', \tau)$ and a coupling between random variables $\gT^{\pi}_\tau Z_1(s, a, \tau)$ and $\gT^{\pi}_\tau Z_2(s, a, \tau)$ to relate the $p$-Wasserstein distance $W_p^{\text{sup}}(\gT^{\pi}_\tau Z_1, \gT^{\pi}_\tau Z_2)$ with the $p$-Wasserstein distance $W_p^{\text{sup}}(Z_1, Z_2)$.

Let $F^{-1}_{Z_1}(z \mid S', A', \tau)$ and $F^{-1}_{Z_2}(z \mid S', A', \tau)$ denote the quantile function (inverse cumulative distribution function) of the random variable $Z_1(S', A', \tau)$ and $Z_2(S', A', \tau)$, respectively.
Let $(C_1(S', A', \tau), C_2(S', A', \tau)) \in \gZ \times \gZ$ be a pair of return random variables. We define $(C_1, C_2)$ by transforming a uniform random variable $U \sim \textsc{Unif}([0, 1])$ using the quantile functions $F^{-1}_{Z_1}$ and $F^{-1}_{Z_2}$ at the next state $S'$ and the next action $A'$:
\begin{align*}
    (C_1(S', A', \tau), \: C_2(S', A', \tau)) \triangleq \left( F_{Z_1}^{-1}(U \mid S', A', \tau), \: F_{Z_2}^{-1}(U \mid S', A', \tau) \right), \: U \sim \textsc{Unif}([0, 1]),
\end{align*}
where $U$ is independent of $(S', A')$. Importantly, the random variables $C_1$ and $C_2$ are generally dependent on each other because they depend on the same set of random variables $(S', A', U)$. By definition, we can show that $(C_1(S', A', \tau), C_2(S', A', \tau))$ is a coupling between $Z_1(S', A', \tau)$ and $Z_2(S', A', \tau)$ by computing their CDFs: for any $y \in \R$,
\begin{align*}
    F_{C_1}(y \mid S', A', \tau) &\overset{(a)}{=} \mathbb{P} \left( F^{-1}_{Z_1}(U \mid S', A', \tau) \leq y \right) \\
    &\overset{(b)}{=} \mathbb{P} \left( U \leq F_{Z_1}(y \mid S', A', \tau) \right) \\
    &\overset{(c)}{=} F_{Z_1}(y \mid S', A', \tau),
\end{align*}
where \emph{(a)} uses the definition of the conditional CDF, \emph{(b)} applies the CDF $F_{Z_1}$ on both sides of the inequality, and \emph{(c)} uses the CDF of the uniform random variable. Therefore, the random variable $C_1$ has the same law as $Z_1$: $C_1(S', A', \tau) \overset{d}{=} Z_1(S', A', \tau)$. Similarly, we can show that $C_2(S', A', \tau) \overset{d}{=} Z_2(S', A', \tau)$. Together, these two equalities in distribution show that the joint distribution $(C_1(S', A', \tau), C_2(S', A', \tau))$ is a coupling between $Z_1(S', A', \tau)$ and $Z_2(S', A', \tau)$. This coupling recovers the one-dimensional quantile formula for the $p$-Wasserstein distance:
\begin{align}
    W_p(Z_1(S', A', \tau), Z_2(S', A', \tau)) &= \left( \int_0^1 \abs {F_{Z_1}^{-1}(u \mid S', A', \tau) - F_{Z_2}^{-1}(u \mid S', A', \tau)}^p du \right)^{1 / p} \nonumber \\
    &= \E_{ U } \left[ \abs{ F_{Z_1}^{-1}(U \mid S', A', \tau) - F_{Z_2}^{-1}(U \mid S', A', \tau) }^p \right]^{1 / p} \nonumber \\
    &\overset{(a)}{=} \E_{ U } \left[ \abs{ C_1(S', A', \tau) - C_2(S', A', \tau) }^p \right]^{1 / p},
    \label{eq:p-wasserstein-c-coupling}
\end{align}
where \emph{(a)} plugs in the definition of random variables $C_1$ and $C_2$.

We next define a joint distribution that couples the random variables $\gT^{\pi}_\tau Z_1(s, a, \tau)$ and $\gT^{\pi}_\tau Z_2(s, a, \tau)$. Specifically, let $(T_1(s, a, \tau), \: T_2(s, a, \tau)) \in \gZ \times \gZ$ be another pair of return random variables. We define $(T_1, T_2)$ as transforming the random variables $C_1$ and $C_2$ using the flow Bellman equation (Eq.~\ref{eq:flow-bellman-eq}) with a Gaussian correction $N \sim \gN(0, 1)$ that is independent of $(S', A', U)$:
\begin{align*}
    (T_1(s, a, \tau), \: T_2(s, a, \tau)) &\triangleq \left( \tau r(s, a) + \gamma C_1(S', A', \tau) + (1 - \tau) \sqrt{1 - \gamma^2} N, \right. \\
    & \hspace{1.6em} \left. \tau r(s, a) + \gamma C_2(S', A', \tau) + (1 - \tau) \sqrt{1 - \gamma^2} N \right), \\
    S' \sim p(S' \mid s, a), \: A' &\sim \pi(A' \mid S'), \: U \sim \textsc{Unif}([0, 1]), \: N \sim \gN(0, 1).
\end{align*}
Here, the random variables $T_1$ and $T_2$ are also generally dependent on each other because they share the same set of random variables $(S', A', U, N)$. Since $(C_1(S', A', \tau), C_2(S', A', \tau))$ is a coupling between $Z_1(S', A', \tau)$ and $Z_2(S', A', \tau)$, we have
\begin{align*}
    T_1(s, a, \tau) &\overset{d}{=} \tau r(s, a) + \gamma C_1(S', A', \tau) + (1 - \tau) \sqrt{1 - \gamma^2} N \\
    &\overset{d}{=} \tau r(s, a) + \gamma Z_1(S', A', \tau) + (1 - \tau) \sqrt{1 - \gamma^2} N \\
    &\overset{d}{=} \gT^{\pi}_\tau Z_1(s, a, \tau),
\end{align*}
which shows that $T_1$ has the same law as $\gT^{\pi}_\tau Z_1$. Similarly, we can show that $T_2(s, a, \tau) \overset{d}{=} \gT^{\pi}_{\tau} Z_2(s, a, \tau)$. Therefore, these two equalities in distribution suggest that the joint distribution $(T_1(s, a, \tau), \: T_2(s, a, \tau))$ is a coupling between random variables $\gT^{\pi}_\tau Z_1(s, a, \tau)$ and $\gT^{\pi}_\tau Z_2(s, a, \tau)$.

Finally, we can show the desired inequality for the supremum of the $p$-Wasserstein distance over all state-action pairs. We consider the $p$-Wasserstein distance between $\gT^{\pi}_\tau Z_1(s, a, \tau)$ and $\gT^{\pi}_\tau Z_2(s, a, \tau)$:
\begin{align*}
    W_p \left( \gT^{\pi}_\tau Z_1(s, a, \tau), \gT^{\pi}_\tau Z_2(s, a, \tau) \right) &\overset{(a)}{\leq} \E_{S', \:A', \:U, \:N} \left[ \abs{T_1(s, a, \tau) - T_2(s, a, \tau)}^p \right]^{1 / p} \\
    &\overset{(b)}{=} \gamma \E_{S', \:A', \: U} \left[ \abs{ C_1(S', A', \tau) - C_2(S', A', \tau) }^p \right]^{1 / p} \\
    &\overset{(c)}{=} \gamma \E_{S', A'} \left[ \left( W_p( Z_1(S', A', \tau), Z_2(S', A', \tau)) \right)^p \right]^{1 / p} \\
    &\overset{(d)}{\leq} \gamma \sup_{s', a'} W_p( Z_1(s', a', \tau), Z_2(s', a', \tau) ) \\
    &\overset{(e)}{=} \gamma W^{\text{sup}}_p(Z_1, Z_2),
\end{align*}
where \emph{(a)} uses the definition of the $p$-Wasserstein distance (the infimum of the $p$-norm over all couplings), \emph{(b)} plugs in the definitions of $T_1$ and $T_2$, \emph{(c)} applies the law of total expectation and Eq.~\ref{eq:p-wasserstein-c-coupling}, \emph{(d)} uses the supremum over the state-action pairs to bound the expectation over the next state and the next action, and \emph{(e)} applies the definition of the supremum of the $p$-Wasserstein distance. Because this inequality holds for every state-action pair $(s, a)$, we can take the supremum over all state-action pairs to conclude that
\begin{align*}
    W^{\text{sup}}_p \left( \gT_\tau^\pi Z_1, \gT_\tau^\pi Z_2 \right) \leq \gamma W^{\text{sup}}_p( Z_1, Z_2 ).
\end{align*}
\end{proof}

\subsection{Theoretical analyses of \methodnamei{}}
\label{appendix:ivf-proofs}

We conduct theoretical analyses to examine the behavior of the expectile-weighted regression objective in Eq.~\ref{eq:value-ret-imp-vel-loss}. Our analyses answer two questions:
\begin{enumerate}[leftmargin=2.5em]
    \item[\textbf{Q1.}] Does the $V$-velocity that minimizes the expectile-weighted loss induce a $V$-estimate that is the expectile of the $Q$-estimate?
    \item[\textbf{Q2.}] Does increasing the expectile recover the optimal value within the support of the dataset?
\end{enumerate}

We proceed by first discussing the setup of optimization and assumptions about the fixed point of the expectile-weighted regression objective. We then show that minimizing the expectile-weighted loss $\gL^\mu$ fits the return velocities of an implicit policy~\citep{hansen2023idql, sikchi2024dual}. Moreover, at the fixed point, the $V$-estimate and the $Q$-estimate induced by the return velocities satisfy the expectile Bellman equations as in~\citet{kostrikov2022offline}. Finally, we show that as the expectile $\mu$ approaches $1$, the induced $V$-estimate converges to the optimal value function within the support of the dataset, as in Theorem 3 of~\citet{kostrikov2022offline}.

\paragraph{Optimization setup.}
We consider an optimization procedure similar to Fitted Q iteration~\citep{ernst2005tree}: at each iteration $k$, we learn the $V$-velocity $v^k_V$ and the $Q$-velocity $v_Q^k$ using the $V$-velocity $v_V^{k - 1}$ and the $Q$-velocity $v_Q^{k - 1}$ from the previous iteration. Here, $k = 1, 2, \cdots$ indexes the iterations, with $k = 0$ denoting random initialization. We consider learning $v^k_V$ \emph{until convergence} and then increase the number of iterations. In practice, we perform one gradient update at each iteration, which preserves a similar global convergence guarantee~\citep{gaur2023global, cai2020neuraltemporaldifferenceqlearningprovably}. The analysis below assumes convergence to a self-consistent fixed point.

Under this setup, we can write the expectile-weighted objective at iteration $k$ as
\begin{align*}
    \gL^\mu_k(v_V) = \E_{ \substack{ (s, \: a) \sim D, \: \tau \sim \textsc{Unif}([0, 1]), \\ z^\tau \sim p^{k - 1}_Q(z^\tau \mid s, a, \tau) } } \left[ w^{\mu}_{k - 1}(s, a) (v_V(z^{\tau} \mid s, \tau) - v_Q^{k - 1}(z^\tau \mid s, a, \tau) )^2 \right],
\end{align*}
where $v_Q^{k - 1}$ is the $Q$-velocity from the previous iteration $k - 1$, which induces the probability density path $p_Q^{k - 1}$, and the expectile weight $w^\mu_{k - 1}$ is defined using the return velocities from the iteration $k - 1$ as
\begin{align}
    Q^{k - 1}(s, a) &\triangleq \E_{z^0 \sim \gN(0, 1)} \left[ \int_0^1 v_Q^{k - 1}(z^\tau \mid s, a, \tau) d \tau \right], \label{eq:q-km1-def} \\
    V^{k - 1}(s) &\triangleq \E_{z^0 \sim \gN(0, 1)} \left[ \int_0^1 v_V^{k - 1}(z^\tau \mid s, \tau) d \tau \right], \label{eq:v-km1-def} \\
    w^{\mu}_{k - 1}(s, a) &\triangleq \mu \mathbbm{1} \left( Q^{k - 1}(s, a) \geq V^{k - 1}(s) \right) + (1 - \mu) \mathbbm{1}( Q^{k - 1}(s, a) < V^{k - 1}(s)). \label{eq:w-km1-def}
\end{align}
Similarly, the objective of the $Q$-velocity at iteration $k$ is
\begin{align}
    \gL_k(v_Q) &= \E_{ \substack{(s, a) \sim D, \: s' \sim p(s' \mid s, a), \\
    \tau \sim \textsc{Unif}([0, 1]), \\
    \epsilon \sim \gN(0, 1), \: z^\tau \sim p^{k - 1}_V(z^{\tau} \mid s', \tau ) }} \left[ \left( v_Q \left( \tau r + \gamma z^\tau + (1 - \tau) \sqrt{1 - \gamma^2} \epsilon \, \middle| \, s, a, \tau \right) \right. \right. \nonumber \\
    &\hspace{12.5em} \left. \left. - \left( r + \gamma v_V^{k - 1}(z^\tau \mid s', \tau) - \sqrt{1 - \gamma^2} \epsilon \right) \right)^2 \right],
    \label{eq:critic-ret-obj-iter-k}
\end{align}
At each iteration $k$, we find the $V$-velocity $v^{k}_V$ by minimizing the expectile-weighted loss $\gL^{\mu}_k(v_V)$ until convergence and find the $Q$-velocity $v_Q^k$ by minimizing the $Q$-velocity loss $\gL_k(v_Q)$ until convergence.

\paragraph{Assumption.} We assume that the iterative optimization procedure above will converge to a fixed point when $k \to \infty$. At such a fixed point, the $V$-velocity $v^{\infty}_V$ minimizes the expectile-weighted loss with the expectile weight defined by itself (self-consistent):
\begin{align}
    v_V^{\infty} &= \argmin_{v_V} \: \gL_{\infty}^\mu(v_V), \nonumber \\
    \gL^\mu_\infty(v_V^{\infty}) &= \E_{ \substack{ (s, a) \sim D, \: \tau \sim \textsc{Unif}([0, 1]), \\ z^\tau \sim p^{\infty}_Q(z^\tau \mid s, a, \tau) } } \left[ w^{\mu}_{\infty}(s, a) (v_V^{\infty}(z^{\tau} \mid s, \tau) - v_Q^{\infty}(z^\tau \mid s, a, \tau) )^2 \right],
    \label{eq:value-vel-assump}
\end{align}
where $w^{\mu}_{\infty}(s, a) = \mu \mathbbm{1} \left( Q^{\infty}(s, a) \geq V^{\infty}(s) \right) + (1 - \mu) \mathbbm{1}( Q^{\infty}(s, a) < V^{\infty}(s))$. Similarly, the $Q$-velocity $v_Q^{\infty}$ minimizes the $Q$-velocity loss:
\begin{align}
    v_Q^{\infty} &= \argmin_{v_Q} \: \gL_{\infty}(v_Q), \nonumber \\
    \gL_{\infty}(v_Q^{\infty}) &= \E_{ \substack{(s, a, r) \sim D, \: s' \sim p(s' \mid s, a), \\
    \tau \sim \textsc{Unif}([0, 1]), \\
    \epsilon \sim \gN(0, 1), \: z^\tau \sim p^{\infty}_V(z^{\tau} \mid s', \tau ) }} \left[ \left( v_Q \left( \tau r + \gamma z^\tau + (1 - \tau) \sqrt{1 - \gamma^2} \epsilon \, \middle| \, s, a, \tau \right) \right. \right. \nonumber \\
    &\hspace{12.5em} \left. \left. - \left( r + \gamma v_V^{\infty}(z^\tau \mid s', \tau) - \sqrt{1 - \gamma^2} \epsilon \right) \right)^2 \right].
    \label{eq:critic-vel-assump}
\end{align}

Under this setup and assumption, we can answer the aforementioned questions.

\textbf{A1: Yes, the fixed point of optimizing the expectile-weighted loss induces an expectile value.} 

We first find the minimizer $v^k_V$ of the bootstrapped flow-matching loss $\gL_k^{\mu}(v_V)$ and the minimizer $v^k_Q$ of the $Q$-velocity loss $\gL_k(v_Q)$ by writing the losses in the forms as in Lemma~\ref{lemma:stochastic-process-flow-matching}. The main idea is to define stochastic processes similar to those in Sec.~\ref{subsec:learning-flow-vels} and employ Lemma~\ref{lemma:stochastic-process-flow-matching}. We then show that finding the fixed point of the bootstrapped flow-matching loss and the $Q$-velocity loss helps us recover the flow Bellman equations (Eq.~\ref{eq:flow-bellman-eqs}) at convergence. In particular, this flow Bellman equation corresponds to an implicit policy. Finally, we use the flow Bellman equation to show that the induced $V$-estimate and $Q$-estimate satisfy the expectile Bellman equations as in~\citet{kostrikov2022offline}.

\paragraph{Minimizer of the $V$-flow loss $\gL^{\mu}_k(v_V)$.} We start by defining an \emph{implicit} policy~\citep{hansen2023idql} at each iteration:
\begin{align}
    \pi_{k - 1}^\mu(a \mid s) \triangleq \frac{ \pi_\beta(a \mid s) w^\mu_{k - 1}(s, a) }{ \E_{\pi_\beta( a \mid s )} [w^\mu_{k - 1}(s, a)] },
    \label{eq:implicit-policy}
\end{align}
where $\pi_\beta(a \mid s)$ is the behavioral policy that collects the actions in the dataset. Since $w_{k - 1}^\mu \in [1 - \mu, \mu]$ and $\mu \in [0.5, 1)$, the implicit policy remains well-defined (non-negative density). Using this implicit policy, we can rewrite the bootstrapped flow-matching loss $\gL_k^{\mu}(v_V)$ as
\begin{align*}
    \gL_k^{\mu}( v_V ) \overset{\text{const.}}{=} \E_{ \substack{s \sim D, \: {\color{mygreen} a \sim \pi^\mu_{k - 1}(a \mid s)} \\ \tau \sim \textsc{Unif}([0, 1]), \\ z^\tau \sim p_Q^{k - 1} \left( z^\tau \mid s, a, \tau \right) }} \left[ \left( v_V(z^\tau \mid s, \tau) - v_Q^{k - 1}( z^\tau \mid s, a, \tau ) \right)^2 \right],
\end{align*}
where $\text{const.}$ denotes the denominator in the implicit policy (Eq.~\ref{eq:implicit-policy}) that does not affect the pointwise minimizer of $\gL_k^{\mu}(v_V)$ with respect to states. Similar to the discussion in Sec.~\ref{subsec:learning-flow-vels}, we can write the sampling of noisy return $z^\tau \sim p^{k - 1}_Q(z^\tau \mid s, a, \tau)$ as transporting a Gaussian noise $z^0 \sim \gN(0, 1)$ using the flow $\phi^{k - 1}_Q(z^0 \mid s, a, \tau)$ induced by the $Q$-velocity $v^{k - 1}_Q(z^\tau \mid s, a, \tau)$:
\begin{align*}
    z^\tau \sim p^{k - 1}_Q(z^\tau \mid s, a, \tau) \iff z^0 \sim \gN(0, 1), \: z^\tau = \phi^{k - 1}_Q(z^0 \mid s, a, \tau),
\end{align*}
and define the $V$-flow stochastic process corresponding to the expectile-weighted objective $\gL_k^{\mu}( v_V )$ as
\begin{align*}
    X_V^k(s, \tau) \triangleq \phi_Q^{k - 1}( N_Q \mid s, A, \tau ), \quad \frac{\de}{\de \tau} X^k_V(s, \tau) \overset{(a)}{=} v^{k - 1}_Q( \phi^{k - 1}_Q(N_Q \mid s, A, \tau) \mid s, A, \tau),
\end{align*}
where $N_Q$ is the Gaussian random variable, $A$ is the action random variable as in Sec.~\ref{subsec:learning-flow-vels}, and \emph{(a)} is true because $v^{k - 1}_Q$ and $\phi^{k - 1}_Q$ satisfy their own flow ODE (Eq.~\ref{eq:flow-ode}). We can also collect the set of random variables for this stochastic process as $\gX_{V}^k = \{ N_Q, A \}$. Applying Lemma~\ref{lemma:stochastic-process-flow-matching} gives that the minimizer $v^k_V$ of the expectile-weighted objective $\gL^{\mu}_k(v_V)$ fits to the noise-level differentiation $\de X^k_V(s, \tau) / \de \tau$:
\begin{align*}
    v_V^k(z^\tau \mid s, \tau) &= \argmin_{v_V} \: \gL^{\mu}_k(v_V) \\
    &= \E_{\gX_V^k} \left[ \frac{\de}{\de \tau} X^k_V(s, \tau) \middle| X^k_V(s, \tau) = z^\tau \right] \\
    &= \E_{{\color{mygreen} a \sim \pi^{\mu}_{k - 1}(a \mid s)}, \: z^\tau \sim p^{k - 1}_Q(z^\tau \mid s, a, \tau) } \left[ v^{k - 1}_Q(z^\tau \mid s, a, \tau) \right].
\end{align*}
In addition, Lemma~\ref{lemma:stochastic-process-flow-matching} suggests that the $V$-velocity $v^k_V(z^\tau \mid s, \tau)$ generates the probability density path of the stochastic process $X^k_V(s, \tau)$, i.e., $X^k_V(s, \tau)$ has the same law as the flow $\phi^k_V$ that transports a Gaussian random variable $N_V$ independent of $N_Q$:
\begin{align*}
    \phi^k_V(N_V \mid s, \tau) \overset{d}{=} X^k_V(s, \tau) = \phi_Q^{k - 1}(N_Q \mid s, A, \tau).
\end{align*}
We note that this identity is similar to the $V$-flow in Eq.~\ref{eq:value-flow}.

\paragraph{Minimizer of the $Q$-flow loss $\gL_k(v_Q)$.} Similarly, we can derive the minimizer $v^k_Q$ of the $Q$-velocity objective $\gL_k(v_Q)$. Because the implicit policy $\pi^\mu_{k - 1}$ does not appear in the loss $\gL_k(v_Q)$, we can make an analogy to the discussion in Sec.~\ref{subsec:learning-flow-vels} directly. Specifically, we write the sampling of noisy return $z^\tau \sim p_V^{k - 1}(z^\tau \mid s', \tau)$ in Eq.~\ref{eq:critic-ret-obj-iter-k} as transporting a Gaussian noise $z^0 \sim \gN(0, 1)$ using the flow $\phi^{k - 1}_V(z^0 \mid s', \tau)$ induced by the $V$-velocity $v^{k - 1}_V(z^\tau \mid s', \tau)$:
\begin{align*}
    z^\tau \sim p^{k - 1}_V(z^\tau \mid s', \tau) \iff z^0 \sim \gN(0, 1), \: z^\tau = \phi^{k - 1}_V(z^0 \mid s', \tau),
\end{align*}
and define the $Q$-return stochastic process corresponding to the $Q$-velocity $\gL_k(v_Q)$ as
\begin{align*}
    X^{k}_Q(s, a, \tau) &\triangleq \tau r(s, a) + \gamma \phi^{k - 1}_V(N_V \mid S', \tau) + (1 - \tau) \sqrt{1 - \gamma^2} N, \\
    \frac{\de}{\de \tau} X^{k}_Q(s, a, \tau) &\overset{(a)}{=} r(s, a) + \gamma v^{k - 1}_V( \phi^{k - 1}_V(N_V \mid S', \tau) \mid S', \tau) - \sqrt{1 - \gamma^2} N,
\end{align*}
where $N_V$ and $N$ are independent Gaussian random variables as in Sec.~\ref{subsec:learning-flow-vels}, $S'$ is the random variable of the next state as in Eq.~\ref{eq:dist-bellman-eq}, and \emph{(a)} is correct because $v^{k - 1}_V$ and $\phi^{k - 1}_V$ satisfy their own flow ODE (Eq.~\ref{eq:flow-ode}). We also collect the set of random variables for this stochastic process as $\gX_Q^k = \{ N_V, N, S' \}$ and apply Lemma~\ref{lemma:stochastic-process-flow-matching} to find the minimizer $v^k_Q$ of the $Q$-velocity objective $\gL_k(v_Q)$:
\begin{footalign*}
    v^k_Q(z^\tau \mid s, a, \tau) &= \argmin_{v_Q} \gL_k(v_Q) \\
    &= \E_{ \gX^k_Q } \left[ \frac{\de }{\de \tau} X^k_Q(s, a, \tau) \middle| X^k_Q(s, a, \tau) = z^\tau \right] \\
    &\overset{(a)}{=} \E_{ \substack{ s' \sim p(s' \mid s, a), \epsilon \sim \gN(0, 1) \\ z^\tau \sim p^{k - 1}_V(z^\tau \mid s, a, \tau) } } \left[ r - \sqrt{1 - \gamma^2} \epsilon \right. \\
    &\left. \hspace{11em} + \gamma v_V^{k - 1} \left( \tau r + \gamma z^\tau + (1 - \tau) \sqrt{1 - \gamma^2} \epsilon \: \middle| \: s', \tau \right) \right],
\end{footalign*}
where we do not scale the reward function in \emph{(a)} by $(1 - \gamma)$ as in Sec.~\ref{subsec:learning-flow-vels} because we are focusing on theoretical analysis.
In addition, Lemma~\ref{lemma:stochastic-process-flow-matching} suggests that the $Q$-velocity $v^k_Q(z^\tau \mid s, a, \tau)$ generates the probability density path of the stochastic process $X^k_Q(s, a, \tau)$, i.e., $X^k_Q(s, a, \tau)$ has the same law as the flow $\phi^k_Q$ that transports a Gaussian random variable $N_Q$ independent of $N_V$ and $N$:
\begin{align*}
    \phi^k_Q(N_Q \mid s, a, \tau) &\overset{d}{=} \tau r(s, a) + \gamma \phi^{k - 1}_V(N_V \mid S', \tau) + (1 - \tau) \sqrt{1 - \gamma^2} N.
\end{align*}
We also note that this identity is similar to the $Q$-flow in Eq.~\ref{eq:critic-flow}.

\paragraph{Flow Bellman equations at convergence.} Therefore, under the assumption in Eq.~\ref{eq:value-vel-assump} and Eq.~\ref{eq:critic-vel-assump} and when $k \to \infty$, the implicit policy can be written as
\begin{align*}
    \pi^\mu_{\infty}(a \mid s) = \frac{ \pi_{\beta}(a \mid s) w_{\infty}^\mu(s, a) }{ \E_{\pi_{\beta}(a \mid s)} \left[ w^\mu_{\infty}(s, a) \right] },
\end{align*}
and we have
\begin{align*}
    \phi_V^{\infty}(N_V \mid s, \tau) &\overset{d}{=} \phi_Q^{\infty}( N_Q \mid s, A, \tau), \\
    \phi_Q^{\infty}(N_Q \mid s, a, \tau) &\overset{d}{=} \tau r(s, a) + \gamma \phi_V^{\infty}( N_V \mid S', \tau) + (1 - \tau) \sqrt{1 - \gamma^2} N,
\end{align*}
which are exactly the flow Bellman equations (Eq.~\ref{eq:flow-bellman-eqs}) of the implicit policy $\pi^\mu_{\infty}$ with a unique fixed point:
\begin{align*}
    \phi^{\infty}_V (N_V \mid s, \tau) \overset{d}{=} Z^{\pi_{\infty}^\mu}_V( s, \tau ), \quad \phi^{\infty}_Q (N_Q \mid s, a, \tau) \overset{d}{=} Z^{\pi_{\infty}^\mu}_Q ( s, a, \tau ).
\end{align*}
As discussed in Sec.~\ref{subsec:flow-bellman-eq}, at the noise level $\tau = 1$, we recover the standard distributional Bellman equations:
\begin{align*}
    \phi^{\infty}_V(N_V \mid s,  1) \overset{d}{=} \phi^{\infty}_Q(N_Q \mid s, A, 1), \quad \phi^{\infty}_Q(N_Q \mid s, a, 1) &\overset{d}{=} r(s, a) + \gamma \phi^{\infty}_V(N_V \mid S', 1).
\end{align*}
\paragraph{Expectile Bellman equations for $V$- and $Q$-estimate.} Taking expectations over the random variables $(N_Q, N, A, S')$ on both sides of these identities, we have
\begin{align*}
    V^{\infty}(s) &\overset{(a)}{=} \E_{z^0 \sim \gN(0, 1)} \left[ \int_0^1 v_V^{\infty}(z^\tau \mid s, \tau) d \tau \right]  \\
    &= \E_{z^0 \sim \gN(0, 1)} \left[ \phi_V^{\infty}(z^0 \mid s, 1) \right] \\
    &\overset{(b)}{=} \E_{ \substack{z^0 \sim \gN(0, 1) \\ {\color{mygreen} a \sim \pi^{\mu}_{\infty}(a \mid s)} } } \left[ \phi_Q^{\infty}(z^0 \mid s, a, 1) \right] \\
    &= \E_{ \substack{z^0 \sim \gN(0, 1) \\ {\color{mygreen} a \sim \pi^{\mu}_{\infty}(a \mid s)} } } \left[ \int_0^1 v_Q^{\infty}(z^\tau \mid s, a, \tau) d\tau \right] \\
    &\overset{(c)}{=} \E_{ {\color{mygreen} a \sim \pi^{\mu}_{\infty}(a \mid s)} } \left[ Q^{\infty}(s, a) \right], \\
    Q^{\infty}(s, a) &\overset{(d)}{=} \E_{z^0 \sim \gN(0, 1)} \left[ \int_0^1 v^{\infty}_Q(z^\tau \mid s, a, \tau) d \tau \right] \\
    &= \E_{z^0 \sim \gN(0, 1)} \left[ \phi_Q^{\infty}(z^0 \mid s, a, 1) \right] \\
    &\overset{(e)}{=} r(s, a) + \gamma \E_{ \substack{ z^0 \sim \gN(0, 1), \\ \: s' \sim p(s' \mid s, a) } } \left[ \phi_V^{\infty}(z^0 \mid s', 1) \right] \\
    &= r(s, a) + \gamma \E_{ \substack{ z^0 \sim \gN(0, 1), \\ \: s' \sim p(s' \mid s, a) }} \left[ \int_0^1 v_V^{\infty}(z^\tau \mid s', \tau) d\tau \right] \\
    &\overset{(f)}{=} r(s, a) + \gamma \E_{ s' \sim p(s' \mid s, a) } \left[ V^{\infty}(s') \right],
\end{align*}
where \emph{(a)} and \emph{(f)} use the definition of $V^{\infty}$ (Eq.~\ref{eq:v-km1-def}), \emph{(b)} and \emph{(e)} apply the identities between $\phi^{\infty}_V$ and $\phi^{\infty}_Q$, and \emph{(c)} and \emph{(d)} use the definition of $Q^{\infty}$ (Eq.~\ref{eq:q-km1-def}). Since the implicit policy $\pi^{\mu}_\infty$ itself is defined by the scalar $V^{\infty}$ and the scalar $Q^{\infty}$, applying Theorem~4.1 and Eq.~9 from~\citet{hansen2023idql}, we conclude that $V^{\infty}$ and $Q^{\infty}$ satisfy the expectile Bellman equations. Because different expectiles $\mu$ induce different $V^{\infty}$ and $Q^{\infty}$, we use $V^{\infty}_\mu$ and $Q^{\infty}_\mu$ to denote $V^{\infty}$ and $Q^{\infty}$ for a fixed $\mu$ ($V_\tau$ and $Q_\tau$ in~\citet{kostrikov2022offline}). Our answers to the first question help us to address the second question easily.

\textbf{A2: Yes, increasing the expectile recovers the optimal value within the support of the dataset.}

We reuse theoretical results from~\citep{kostrikov2022offline} to show the answer. For a fixed expectile $\mu$, when the number of iterations $k \to \infty$, we have shown that the scalar $V^{\infty}_\mu$ and the scalar $Q^{\infty}_{\mu}$ induced by their velocities satisfy the expectile Bellman equations: for any $(s, a)$,
\begin{align*}
    V^{\infty}_{\mu}(s) = \E_{ {\color{mygreen} a \sim \pi^{\mu}_{\infty}(a \mid s)} } \left[ Q^{\infty}_{\mu}(s, a) \right], \quad 
    Q^{\infty}_{\mu}(s, a) = r(s, a) + \gamma \E_{s' \sim p(s' \mid s, a)} \left[ V^{\infty}_{\mu}(s') \right].
\end{align*}
Applying Theorem 3 from~\citet{kostrikov2022offline} yields that, as the expectile $\mu$ reaches $1$,
\begin{align*}
    \lim_{\mu \to 1} V^{\infty}_{\mu}(s) &= \max_{a: \: \pi_\beta(a \mid s) > 0} Q^{\star}_{\beta}(s, a) = V^{\star}_{\beta}(s), \\
    \lim_{\mu \to 1} Q^{\infty}_{\mu}(s, a) &= r(s, a) + \gamma \E_{s' \sim p(s' \mid s, a)} \left[ V^{\star}_\beta(s') \right] = Q^{\star}_\beta(s, a),
\end{align*}
where $V^{\star}_\beta$ and $Q^{\star}_\beta$ are the optimal value and critic within the support of the dataset: $a \in \gA \: \text{s.t.} \: \pi_\beta(a \mid s) > 0$. Therefore, we conclude that, when increasing the expectile $\mu$ to $1$, the scalar $V^{\infty}_{\mu}$ and the scalar $Q^{\infty}_{\mu}$ recover the optimal value $V^{\star}_\beta$ and the optimal critic $Q^{\star}_\beta$ within the support of the dataset.

\section{Additional Experiments}

\begin{figure*}[t]
    \centering
    \includegraphics[width=\linewidth]{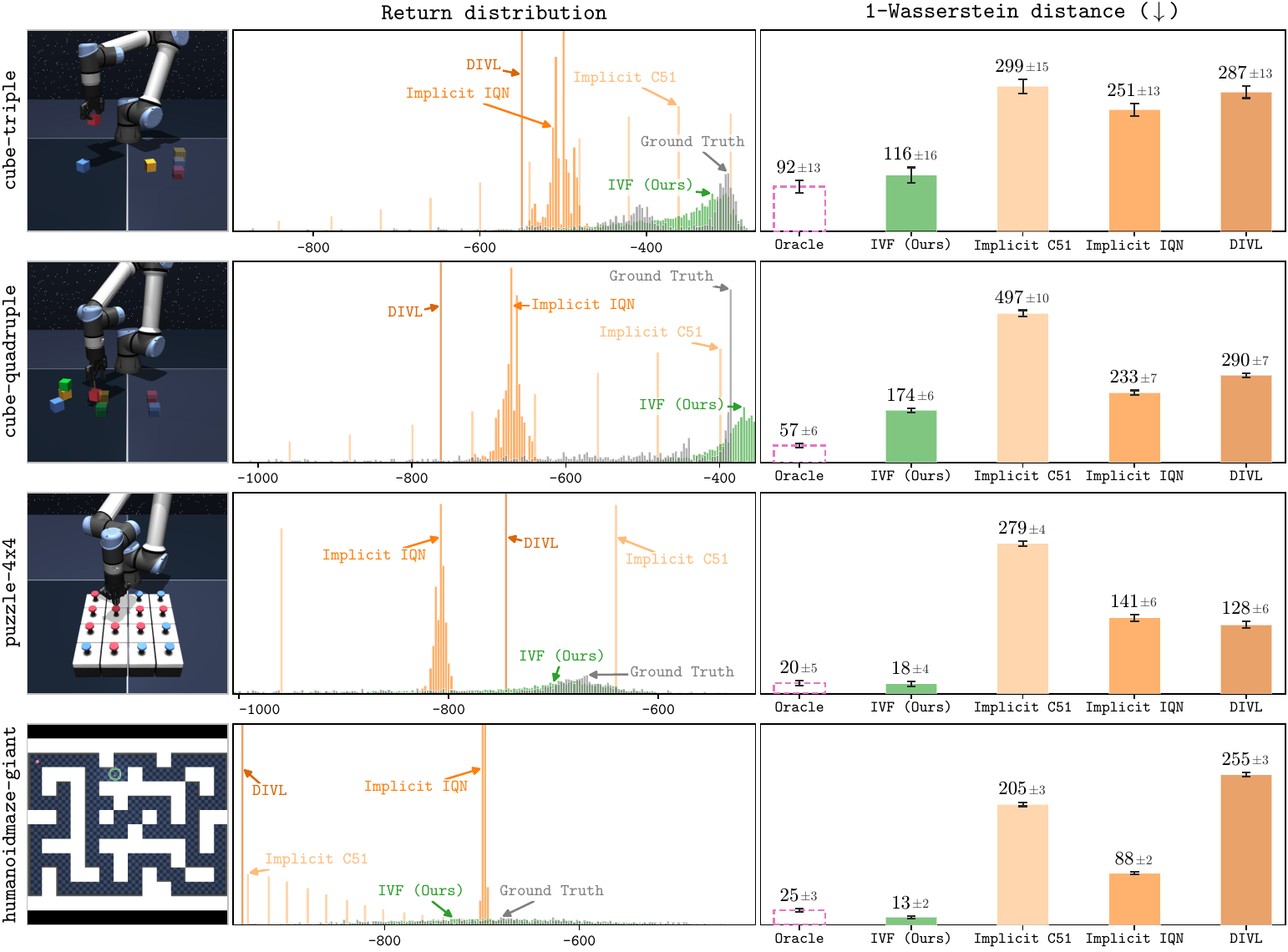}
    \caption{
    Similar to the results in Sec.~\ref{subsec:ret-hist}, \methodnamei{} learns more accurate return distributions than implicit distributional RL baselines. 
    \figleft{} We conduct policy evaluation experiments on the same set of states from the four OGBench domains. \figcenter{} \methodnamei{} better captures the ground-truth return distributions, while Implicit C51, Implicit IQN, and DIVL produce shifted or overly concentrated predictions. \figright{} \methodnamei{} achieves lower mean $1$-Wasserstein distances than all three baselines at the evaluated state in each of the $4$ domains.
    }
    \label{fig:implicit-ret-hist}
\end{figure*}

\subsection{Visualizing return flows across noise levels}
\label{appendix:ret-flow-paths}

\begin{figure*}[t]
    \centering
    \includegraphics[width=\linewidth]{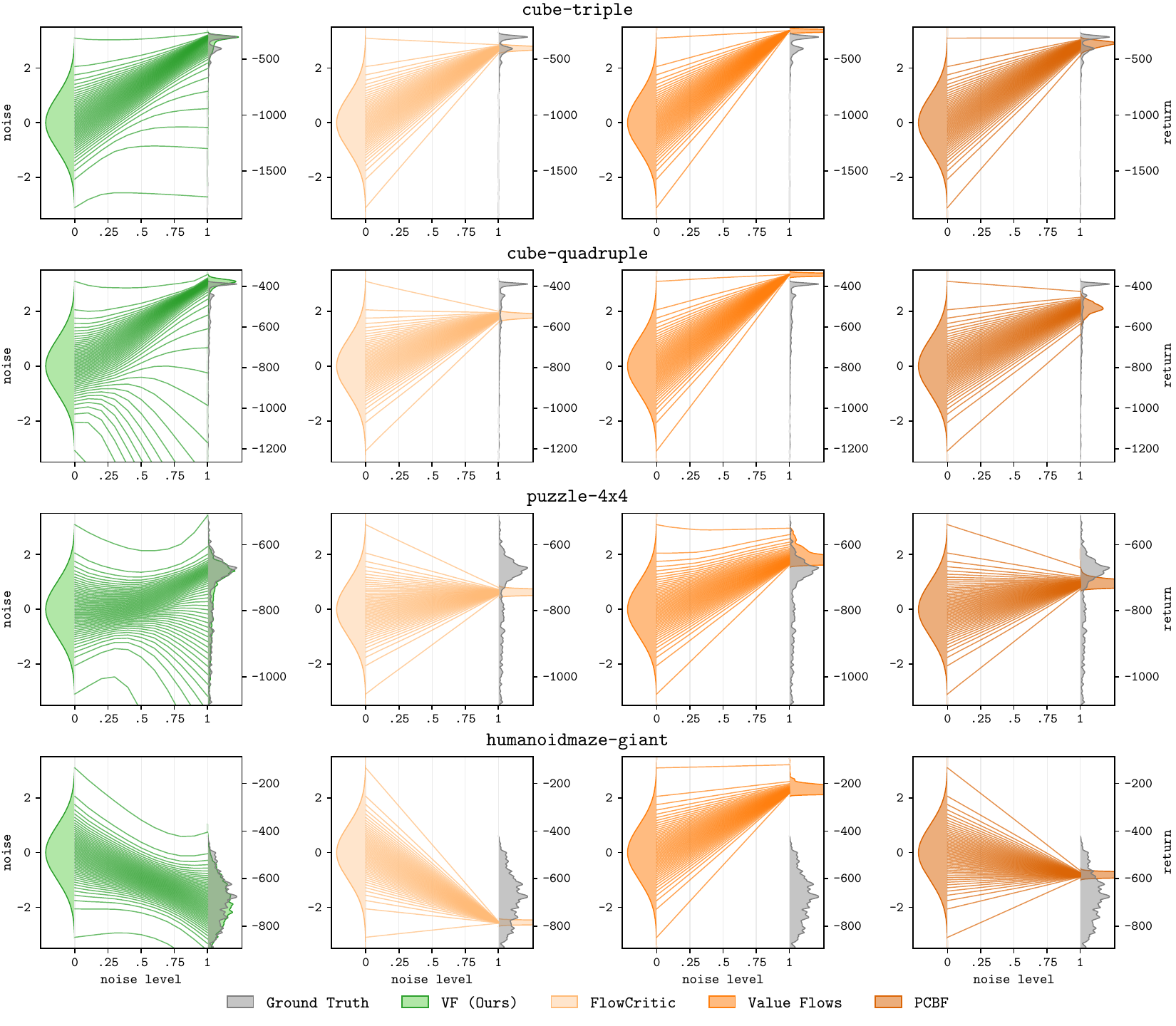}
    \caption{\footnotesize\textbf{\methodexplain{} learns return flows that better capture the spread of ground-truth returns.} 
    Following the policy evaluation experiments in Sec.~\ref{subsec:ret-hist}, we compare the return flow paths of \methodname{} (\emph{Col 1}), FlowCritic (\emph{Col 2}), Value Flows (\emph{Col 3}), and PCBF (\emph{Col 4}) from Gaussian noise ($\tau=0$) to predicted returns ($\tau=1$). \methodname{} spreads its return paths at early noise levels (around $\tau = 0.25$), while baselines concentrate their return paths close to the flow endpoint (around $\tau = 0.9$).
    }
    \label{fig:ret-flows}
\end{figure*}

We next examine how the learned flows transport Gaussian noise into return predictions. Following the policy evaluation setup in Sec.~\ref{subsec:ret-hist}, we compare \methodname{}, FlowCritic, Value Flows, and PCBF at the same state from each domain, with a fixed starting action chunk. For each method, we sample initial noise $z^0 \sim \gN(0,1)$ and numerically integrate the learned $Q$-velocities from $\tau = 0$ to $\tau = 1$ with $10$ Euler steps, recording the intermediate noisy return $z^\tau$. We draw the return flow paths of $51$ return samples starting from the $0$th, $2$nd, $4$th, \dots, $100$th quantiles of the Gaussian distribution. We also plot the initial noise density and the predicted return density, together with the ground-truth return density at the endpoint. We show return predictions that span from the $5$th percentile of the ground-truth returns to the largest return, extended to cover every Value Flows prediction.

Fig.~\ref{fig:ret-flows} shows the evolutions of return flow paths. Although all methods start from Gaussian noise, \methodname{} preserves flow paths that lead to different regions of the return distribution. FlowCritic and Value Flows instead concentrate many flow paths toward a common return value as the noise level approaches $1$. PCBF produces broader predictions on the \texttt{cube} domains, but its endpoint distributions remain overly concentrated, particularly on \texttt{puzzle-4x4} and \texttt{humanoidmaze-giant}.
In addition, different methods also differ in where their trajectories terminate. On \texttt{cube-quadruple} and \texttt{humanoidmaze-giant}, FlowCritic underestimates the return distribution, while Value Flows overestimates it. PCBF places its predictions closer to the ground-truth peak on \texttt{humanoidmaze-giant}, but still misses much of the return spread. \methodname{} captures both the location and spread of the ground-truth return distributions. Together with the $1$-Wasserstein distance comparisons in Fig.~\ref{fig:ret-hist}, these visualizations suggest that noisy return alignment helps spread the return samples while improving the accuracy of the final predictions.

\subsection{Additional ablations on return flow hyperparameters}
\label{appendix:hyperparam-ablations}

In this section, we study how the two important hyperparameters in return flow affect policy performance for both \methodname{} and \methodnamei{}. Because of compute constraints, we ablate these parameters on \texttt{cube-triple} and \texttt{puzzle-4x4} with task $2$ and task $4$, which are representative tasks in those domains.

\begin{figure*}[t]
    \centering
    \begin{subfigure}[c]{0.6\textwidth}
        \centering
        \includegraphics[width=\linewidth]{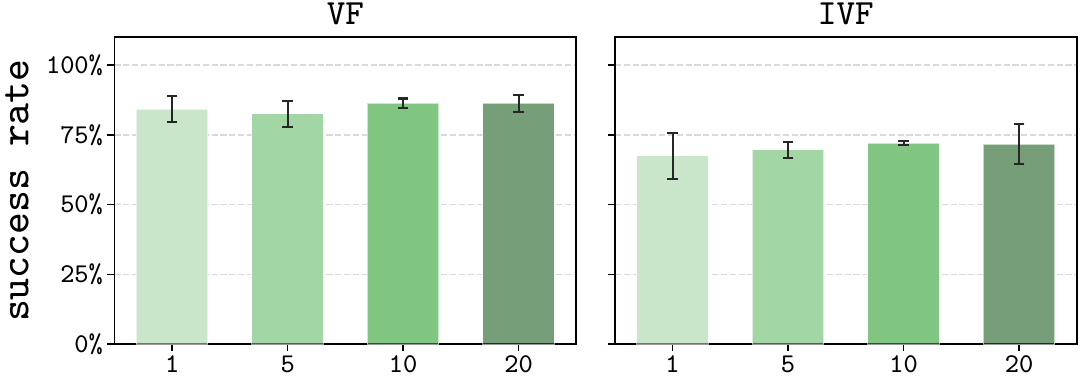}
        \caption{Number of return flow steps during ODE integration.}
        \label{fig:hyperparam-ablation-ret-flow-steps}
    \end{subfigure}
    \hfill
    \begin{subfigure}[c]{0.6\textwidth}
        \centering
        \includegraphics[width=\linewidth]{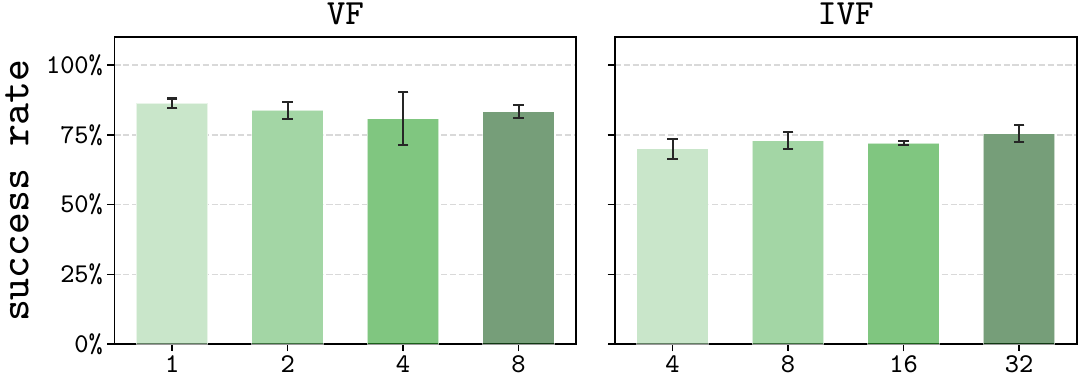}
        \caption{Number of return samples $M$ for estimating $V$ and $Q$}
        \label{fig:hyperparam-ablation-num-ret-samples}
    \end{subfigure}
    \caption{ \textbf{Return flow hyperparameter ablations.} We conduct ablation experiments to study the sensitivity of hyperparameters related to return flows on task $2$ and task $4$ of \texttt{cube-triple} and \texttt{puzzle-4x4}. For each hyperparameter, we show the success rates after aggregating over these $4$ tasks.
    }
    \label{fig:hyperparam-ablations}
\end{figure*}

First, we ablate the number of flow steps used for the return flows $v_V$ and $v_Q$. Fig.~\ref{fig:hyperparam-ablation-ret-flow-steps} shows that different numbers of flow steps have no significant impact on policy performance for both \methodexplain{} algorithms. Perhaps surprisingly, $1$ flow step can yield similar performance.

Second, we ablate the number of return samples $M$ for estimating $V$ and $Q$ (see Sec.~\ref{sec:value-forcing-algorithms}). Fig.~\ref{fig:hyperparam-ablation-num-ret-samples} shows that different numbers of $M$ do not have a significant impact.

\section{Complete numerical results}
\label{appendix:complete-results}

For completeness, Tables~\ref{tab:full-results-explicit-10}-\ref{tab:full-results-implicit-100} provide the full results for every evaluated task and dataset size. The results are averaged over $4$ seeds, and we report the mean and the standard deviations for each task. We highlight the results at or above $95\%$ of the best mean performance within each method group using {\color{mygreen} green}.

\begin{table*}[t]
\centering
\caption{\footnotesize Complete results for explicit and implicit methods with $10\%$ of the dataset. We highlight nonzero mean performance at or above $95\%$ of the best in each row within each method group in {\color{mygreen}green}.}
\label{tab:full-results-10}
\label{tab:full-results-explicit-10}
\label{tab:full-results-implicit-10}
\vspace{-0.5em}
\begingroup
\small
\setlength{\tabcolsep}{2pt}
\renewcommand{\arraystretch}{1.10}
\resizebox{\textwidth}{!}{%
\begin{tabular}{@{}l *{8}{r@{\hspace{0.05cm}}l}@{\hspace{0.18cm}} *{5}{r@{\hspace{0.05cm}}l}@{}}
\toprule
 & \multicolumn{16}{c}{\textit{Explicit methods}} & \multicolumn{10}{c}{\textit{Implicit methods}} \\
\cmidrule(lr){2-17}\cmidrule(l){18-27}
\texttt{Environment} & \hphantom{\texttt{00}}\makebox[0pt]{\texttt{Floq}} & \hphantom{\texttt{00}} & \hphantom{\texttt{00000}}\makebox[0pt]{\texttt{FlowCritic}} & \hphantom{\texttt{00000}} & \hphantom{\texttt{000}}\makebox[0pt]{\shortstack{\texttt{Value}\\\texttt{Flows}}} & \hphantom{\texttt{000}} & \hphantom{\texttt{00}}\makebox[0pt]{\texttt{PCBF}} & \hphantom{\texttt{00}} & \hphantom{\texttt{00}}\makebox[0pt]{\texttt{C51}} & \hphantom{\texttt{00}} & \hphantom{\texttt{00}}\makebox[0pt]{\texttt{IQN}} & \hphantom{\texttt{00}} & \hphantom{\texttt{0}}\makebox[0pt]{\texttt{QC}} & \hphantom{\texttt{0}} & \hphantom{\texttt{000}}\makebox[0pt]{\shortstack{\textcolor{mygreen}{\texttt{VF}}\\\textcolor{mygreen}{\texttt{(Ours)}}}} & \hphantom{\texttt{000}} & \hphantom{\texttt{00}}\makebox[0pt]{\texttt{DIVL}} & \hphantom{\texttt{00}} & \hphantom{\texttt{0000}}\makebox[0pt]{\shortstack{\texttt{Implicit}\\\texttt{C51}}} & \hphantom{\texttt{0000}} & \hphantom{\texttt{0000}}\makebox[0pt]{\shortstack{\texttt{Implicit}\\\texttt{IQN}}} & \hphantom{\texttt{0000}} & \hphantom{\texttt{00}}\makebox[0pt]{\texttt{Implicit QC}} & \hphantom{\texttt{00}} & \hphantom{\texttt{000}}\makebox[0pt]{\shortstack{\textcolor{mygreen}{\texttt{IVF}}\\\textcolor{mygreen}{\texttt{(Ours)}}}} & \hphantom{\texttt{000}} \\
\midrule
\texttt{cube-triple-task1} & \texttt{0} & {\scriptsize $\pm$\texttt{0}} & \texttt{88} & {\scriptsize $\pm$\texttt{19}} & \texttt{70} & {\scriptsize $\pm$\texttt{24}} & \texttt{78} & {\scriptsize $\pm$\texttt{5}} & \texttt{80} & {\scriptsize $\pm$\texttt{22}} & \textcolor{mygreen}{\texttt{100}} & {\scriptsize $\pm$\texttt{0}} & \textcolor{mygreen}{\texttt{95}} & {\scriptsize $\pm$\texttt{10}} & \textcolor{mygreen}{\texttt{100}} & {\scriptsize $\pm$\texttt{0}} & \texttt{10} & {\scriptsize $\pm$\texttt{20}} & \texttt{83} & {\scriptsize $\pm$\texttt{22}} & \texttt{73} & {\scriptsize $\pm$\texttt{28}} & \texttt{75} & {\scriptsize $\pm$\texttt{37}} & \textcolor{mygreen}{\texttt{100}} & {\scriptsize $\pm$\texttt{0}} \\
\texttt{cube-triple-task2} & \texttt{0} & {\scriptsize $\pm$\texttt{0}} & \texttt{68} & {\scriptsize $\pm$\texttt{34}} & \texttt{63} & {\scriptsize $\pm$\texttt{29}} & \texttt{48} & {\scriptsize $\pm$\texttt{21}} & \texttt{58} & {\scriptsize $\pm$\texttt{39}} & \texttt{73} & {\scriptsize $\pm$\texttt{13}} & \texttt{48} & {\scriptsize $\pm$\texttt{13}} & \textcolor{mygreen}{\texttt{90}} & {\scriptsize $\pm$\texttt{8}} & \texttt{0} & {\scriptsize $\pm$\texttt{0}} & \texttt{10} & {\scriptsize $\pm$\texttt{0}} & \texttt{30} & {\scriptsize $\pm$\texttt{29}} & \texttt{48} & {\scriptsize $\pm$\texttt{24}} & \textcolor{mygreen}{\texttt{78}} & {\scriptsize $\pm$\texttt{33}} \\
\texttt{cube-triple-task3} & \texttt{0} & {\scriptsize $\pm$\texttt{0}} & \texttt{83} & {\scriptsize $\pm$\texttt{13}} & \texttt{63} & {\scriptsize $\pm$\texttt{21}} & \texttt{35} & {\scriptsize $\pm$\texttt{25}} & \texttt{43} & {\scriptsize $\pm$\texttt{34}} & \textcolor{mygreen}{\texttt{90}} & {\scriptsize $\pm$\texttt{12}} & \texttt{63} & {\scriptsize $\pm$\texttt{5}} & \texttt{83} & {\scriptsize $\pm$\texttt{15}} & \texttt{0} & {\scriptsize $\pm$\texttt{0}} & \texttt{35} & {\scriptsize $\pm$\texttt{6}} & \texttt{23} & {\scriptsize $\pm$\texttt{10}} & \texttt{35} & {\scriptsize $\pm$\texttt{13}} & \textcolor{mygreen}{\texttt{58}} & {\scriptsize $\pm$\texttt{21}} \\
\texttt{cube-triple-task4} & \texttt{0} & {\scriptsize $\pm$\texttt{0}} & \texttt{23} & {\scriptsize $\pm$\texttt{15}} & \texttt{3} & {\scriptsize $\pm$\texttt{5}} & \texttt{18} & {\scriptsize $\pm$\texttt{5}} & \texttt{23} & {\scriptsize $\pm$\texttt{26}} & \textcolor{mygreen}{\texttt{43}} & {\scriptsize $\pm$\texttt{35}} & \texttt{25} & {\scriptsize $\pm$\texttt{13}} & \texttt{33} & {\scriptsize $\pm$\texttt{26}} & \texttt{0} & {\scriptsize $\pm$\texttt{0}} & \texttt{3} & {\scriptsize $\pm$\texttt{5}} & \texttt{5} & {\scriptsize $\pm$\texttt{10}} & \texttt{8} & {\scriptsize $\pm$\texttt{5}} & \textcolor{mygreen}{\texttt{20}} & {\scriptsize $\pm$\texttt{8}} \\
\texttt{cube-triple-task5} & \texttt{0} & {\scriptsize $\pm$\texttt{0}} & \textcolor{mygreen}{\texttt{83}} & {\scriptsize $\pm$\texttt{10}} & \texttt{48} & {\scriptsize $\pm$\texttt{55}} & \texttt{18} & {\scriptsize $\pm$\texttt{21}} & \texttt{33} & {\scriptsize $\pm$\texttt{36}} & \texttt{53} & {\scriptsize $\pm$\texttt{33}} & \texttt{65} & {\scriptsize $\pm$\texttt{17}} & \textcolor{mygreen}{\texttt{85}} & {\scriptsize $\pm$\texttt{17}} & \texttt{0} & {\scriptsize $\pm$\texttt{0}} & \texttt{23} & {\scriptsize $\pm$\texttt{15}} & \texttt{28} & {\scriptsize $\pm$\texttt{21}} & \texttt{15} & {\scriptsize $\pm$\texttt{10}} & \textcolor{mygreen}{\texttt{70}} & {\scriptsize $\pm$\texttt{32}} \\
\addlinespace[0.2em]
\textit{Domain average} & \texttt{0} & {\scriptsize $\pm$\texttt{0}} & \texttt{69} & {\scriptsize $\pm$\texttt{9}} & \texttt{49} & {\scriptsize $\pm$\texttt{8}} & \texttt{39} & {\scriptsize $\pm$\texttt{5}} & \texttt{47} & {\scriptsize $\pm$\texttt{16}} & \texttt{72} & {\scriptsize $\pm$\texttt{8}} & \texttt{59} & {\scriptsize $\pm$\texttt{5}} & \textcolor{mygreen}{\texttt{78}} & {\scriptsize $\pm$\texttt{10}} & \texttt{2} & {\scriptsize $\pm$\texttt{4}} & \texttt{31} & {\scriptsize $\pm$\texttt{5}} & \texttt{32} & {\scriptsize $\pm$\texttt{10}} & \texttt{36} & {\scriptsize $\pm$\texttt{4}} & \textcolor{mygreen}{\texttt{65}} & {\scriptsize $\pm$\texttt{4}} \\
\midrule
\texttt{cube-quadruple-task1} & \texttt{0} & {\scriptsize $\pm$\texttt{0}} & \texttt{8} & {\scriptsize $\pm$\texttt{10}} & \texttt{8} & {\scriptsize $\pm$\texttt{10}} & \textcolor{mygreen}{\texttt{45}} & {\scriptsize $\pm$\texttt{26}} & \texttt{18} & {\scriptsize $\pm$\texttt{35}} & \texttt{25} & {\scriptsize $\pm$\texttt{29}} & \textcolor{mygreen}{\texttt{45}} & {\scriptsize $\pm$\texttt{31}} & \texttt{28} & {\scriptsize $\pm$\texttt{21}} & \texttt{0} & {\scriptsize $\pm$\texttt{0}} & \texttt{5} & {\scriptsize $\pm$\texttt{6}} & \textcolor{mygreen}{\texttt{18}} & {\scriptsize $\pm$\texttt{35}} & \texttt{10} & {\scriptsize $\pm$\texttt{14}} & \textcolor{mygreen}{\texttt{18}} & {\scriptsize $\pm$\texttt{21}} \\
\texttt{cube-quadruple-task2} & \texttt{0} & {\scriptsize $\pm$\texttt{0}} & \texttt{3} & {\scriptsize $\pm$\texttt{5}} & \texttt{5} & {\scriptsize $\pm$\texttt{6}} & \texttt{25} & {\scriptsize $\pm$\texttt{19}} & \texttt{20} & {\scriptsize $\pm$\texttt{8}} & \texttt{5} & {\scriptsize $\pm$\texttt{6}} & \texttt{15} & {\scriptsize $\pm$\texttt{13}} & \textcolor{mygreen}{\texttt{48}} & {\scriptsize $\pm$\texttt{22}} & \texttt{0} & {\scriptsize $\pm$\texttt{0}} & \texttt{5} & {\scriptsize $\pm$\texttt{6}} & \textcolor{mygreen}{\texttt{8}} & {\scriptsize $\pm$\texttt{5}} & \texttt{3} & {\scriptsize $\pm$\texttt{5}} & \textcolor{mygreen}{\texttt{8}} & {\scriptsize $\pm$\texttt{5}} \\
\texttt{cube-quadruple-task3} & \texttt{0} & {\scriptsize $\pm$\texttt{0}} & \texttt{3} & {\scriptsize $\pm$\texttt{5}} & \texttt{3} & {\scriptsize $\pm$\texttt{5}} & \texttt{0} & {\scriptsize $\pm$\texttt{0}} & \texttt{5} & {\scriptsize $\pm$\texttt{6}} & \texttt{0} & {\scriptsize $\pm$\texttt{0}} & \texttt{8} & {\scriptsize $\pm$\texttt{15}} & \textcolor{mygreen}{\texttt{10}} & {\scriptsize $\pm$\texttt{12}} & \texttt{0} & {\scriptsize $\pm$\texttt{0}} & \texttt{0} & {\scriptsize $\pm$\texttt{0}} & \texttt{0} & {\scriptsize $\pm$\texttt{0}} & \texttt{0} & {\scriptsize $\pm$\texttt{0}} & \texttt{0} & {\scriptsize $\pm$\texttt{0}} \\
\texttt{cube-quadruple-task4} & \texttt{0} & {\scriptsize $\pm$\texttt{0}} & \texttt{0} & {\scriptsize $\pm$\texttt{0}} & \texttt{0} & {\scriptsize $\pm$\texttt{0}} & \texttt{0} & {\scriptsize $\pm$\texttt{0}} & \texttt{0} & {\scriptsize $\pm$\texttt{0}} & \texttt{0} & {\scriptsize $\pm$\texttt{0}} & \texttt{0} & {\scriptsize $\pm$\texttt{0}} & \texttt{0} & {\scriptsize $\pm$\texttt{0}} & \texttt{0} & {\scriptsize $\pm$\texttt{0}} & \texttt{0} & {\scriptsize $\pm$\texttt{0}} & \texttt{0} & {\scriptsize $\pm$\texttt{0}} & \texttt{0} & {\scriptsize $\pm$\texttt{0}} & \texttt{0} & {\scriptsize $\pm$\texttt{0}} \\
\texttt{cube-quadruple-task5} & \texttt{0} & {\scriptsize $\pm$\texttt{0}} & \texttt{0} & {\scriptsize $\pm$\texttt{0}} & \texttt{0} & {\scriptsize $\pm$\texttt{0}} & \texttt{0} & {\scriptsize $\pm$\texttt{0}} & \texttt{3} & {\scriptsize $\pm$\texttt{5}} & \texttt{0} & {\scriptsize $\pm$\texttt{0}} & \textcolor{mygreen}{\texttt{5}} & {\scriptsize $\pm$\texttt{10}} & \texttt{0} & {\scriptsize $\pm$\texttt{0}} & \texttt{0} & {\scriptsize $\pm$\texttt{0}} & \texttt{0} & {\scriptsize $\pm$\texttt{0}} & \texttt{0} & {\scriptsize $\pm$\texttt{0}} & \texttt{0} & {\scriptsize $\pm$\texttt{0}} & \texttt{0} & {\scriptsize $\pm$\texttt{0}} \\
\addlinespace[0.2em]
\textit{Domain average} & \texttt{0} & {\scriptsize $\pm$\texttt{0}} & \texttt{3} & {\scriptsize $\pm$\texttt{2}} & \texttt{3} & {\scriptsize $\pm$\texttt{3}} & \texttt{14} & {\scriptsize $\pm$\texttt{7}} & \texttt{9} & {\scriptsize $\pm$\texttt{8}} & \texttt{6} & {\scriptsize $\pm$\texttt{6}} & \texttt{15} & {\scriptsize $\pm$\texttt{8}} & \textcolor{mygreen}{\texttt{17}} & {\scriptsize $\pm$\texttt{8}} & \texttt{0} & {\scriptsize $\pm$\texttt{0}} & \texttt{2} & {\scriptsize $\pm$\texttt{2}} & \textcolor{mygreen}{\texttt{5}} & {\scriptsize $\pm$\texttt{7}} & \texttt{3} & {\scriptsize $\pm$\texttt{3}} & \textcolor{mygreen}{\texttt{5}} & {\scriptsize $\pm$\texttt{3}} \\
\midrule
\texttt{puzzle-4x4-task1} & \texttt{88} & {\scriptsize $\pm$\texttt{15}} & \texttt{93} & {\scriptsize $\pm$\texttt{10}} & \textcolor{mygreen}{\texttt{100}} & {\scriptsize $\pm$\texttt{0}} & \texttt{93} & {\scriptsize $\pm$\texttt{10}} & \textcolor{mygreen}{\texttt{100}} & {\scriptsize $\pm$\texttt{0}} & \textcolor{mygreen}{\texttt{98}} & {\scriptsize $\pm$\texttt{5}} & \textcolor{mygreen}{\texttt{100}} & {\scriptsize $\pm$\texttt{0}} & \textcolor{mygreen}{\texttt{100}} & {\scriptsize $\pm$\texttt{0}} & \texttt{80} & {\scriptsize $\pm$\texttt{8}} & \texttt{73} & {\scriptsize $\pm$\texttt{17}} & \texttt{70} & {\scriptsize $\pm$\texttt{8}} & \texttt{60} & {\scriptsize $\pm$\texttt{14}} & \textcolor{mygreen}{\texttt{88}} & {\scriptsize $\pm$\texttt{10}} \\
\texttt{puzzle-4x4-task2} & \texttt{25} & {\scriptsize $\pm$\texttt{24}} & \texttt{35} & {\scriptsize $\pm$\texttt{31}} & \texttt{90} & {\scriptsize $\pm$\texttt{8}} & \texttt{33} & {\scriptsize $\pm$\texttt{22}} & \textcolor{mygreen}{\texttt{100}} & {\scriptsize $\pm$\texttt{0}} & \texttt{80} & {\scriptsize $\pm$\texttt{27}} & \textcolor{mygreen}{\texttt{100}} & {\scriptsize $\pm$\texttt{0}} & \textcolor{mygreen}{\texttt{100}} & {\scriptsize $\pm$\texttt{0}} & \texttt{20} & {\scriptsize $\pm$\texttt{14}} & \texttt{15} & {\scriptsize $\pm$\texttt{13}} & \texttt{23} & {\scriptsize $\pm$\texttt{17}} & \texttt{18} & {\scriptsize $\pm$\texttt{5}} & \textcolor{mygreen}{\texttt{38}} & {\scriptsize $\pm$\texttt{19}} \\
\texttt{puzzle-4x4-task3} & \texttt{80} & {\scriptsize $\pm$\texttt{16}} & \texttt{93} & {\scriptsize $\pm$\texttt{10}} & \textcolor{mygreen}{\texttt{95}} & {\scriptsize $\pm$\texttt{10}} & \texttt{83} & {\scriptsize $\pm$\texttt{10}} & \textcolor{mygreen}{\texttt{100}} & {\scriptsize $\pm$\texttt{0}} & \texttt{88} & {\scriptsize $\pm$\texttt{15}} & \textcolor{mygreen}{\texttt{100}} & {\scriptsize $\pm$\texttt{0}} & \textcolor{mygreen}{\texttt{100}} & {\scriptsize $\pm$\texttt{0}} & \texttt{70} & {\scriptsize $\pm$\texttt{12}} & \textcolor{mygreen}{\texttt{80}} & {\scriptsize $\pm$\texttt{14}} & \textcolor{mygreen}{\texttt{78}} & {\scriptsize $\pm$\texttt{13}} & \texttt{63} & {\scriptsize $\pm$\texttt{13}} & \textcolor{mygreen}{\texttt{80}} & {\scriptsize $\pm$\texttt{22}} \\
\texttt{puzzle-4x4-task4} & \texttt{38} & {\scriptsize $\pm$\texttt{22}} & \texttt{55} & {\scriptsize $\pm$\texttt{19}} & \texttt{80} & {\scriptsize $\pm$\texttt{18}} & \texttt{48} & {\scriptsize $\pm$\texttt{10}} & \textcolor{mygreen}{\texttt{100}} & {\scriptsize $\pm$\texttt{0}} & \texttt{60} & {\scriptsize $\pm$\texttt{29}} & \textcolor{mygreen}{\texttt{100}} & {\scriptsize $\pm$\texttt{0}} & \textcolor{mygreen}{\texttt{100}} & {\scriptsize $\pm$\texttt{0}} & \textcolor{mygreen}{\texttt{58}} & {\scriptsize $\pm$\texttt{10}} & \texttt{33} & {\scriptsize $\pm$\texttt{19}} & \texttt{35} & {\scriptsize $\pm$\texttt{13}} & \texttt{30} & {\scriptsize $\pm$\texttt{24}} & \texttt{53} & {\scriptsize $\pm$\texttt{25}} \\
\texttt{puzzle-4x4-task5} & \texttt{23} & {\scriptsize $\pm$\texttt{15}} & \texttt{43} & {\scriptsize $\pm$\texttt{22}} & \texttt{88} & {\scriptsize $\pm$\texttt{5}} & \texttt{48} & {\scriptsize $\pm$\texttt{15}} & \textcolor{mygreen}{\texttt{100}} & {\scriptsize $\pm$\texttt{0}} & \texttt{55} & {\scriptsize $\pm$\texttt{21}} & \textcolor{mygreen}{\texttt{100}} & {\scriptsize $\pm$\texttt{0}} & \texttt{80} & {\scriptsize $\pm$\texttt{18}} & \textcolor{mygreen}{\texttt{48}} & {\scriptsize $\pm$\texttt{13}} & \texttt{18} & {\scriptsize $\pm$\texttt{17}} & \texttt{20} & {\scriptsize $\pm$\texttt{16}} & \texttt{20} & {\scriptsize $\pm$\texttt{14}} & \texttt{45} & {\scriptsize $\pm$\texttt{13}} \\
\addlinespace[0.2em]
\textit{Domain average} & \texttt{51} & {\scriptsize $\pm$\texttt{8}} & \texttt{64} & {\scriptsize $\pm$\texttt{5}} & \texttt{91} & {\scriptsize $\pm$\texttt{4}} & \texttt{61} & {\scriptsize $\pm$\texttt{4}} & \textcolor{mygreen}{\texttt{100}} & {\scriptsize $\pm$\texttt{0}} & \texttt{76} & {\scriptsize $\pm$\texttt{13}} & \textcolor{mygreen}{\texttt{100}} & {\scriptsize $\pm$\texttt{0}} & \textcolor{mygreen}{\texttt{96}} & {\scriptsize $\pm$\texttt{4}} & \texttt{55} & {\scriptsize $\pm$\texttt{5}} & \texttt{44} & {\scriptsize $\pm$\texttt{4}} & \texttt{45} & {\scriptsize $\pm$\texttt{5}} & \texttt{38} & {\scriptsize $\pm$\texttt{3}} & \textcolor{mygreen}{\texttt{61}} & {\scriptsize $\pm$\texttt{7}} \\
\midrule
\texttt{humanoidmaze-giant-task1} & \texttt{0} & {\scriptsize $\pm$\texttt{0}} & \texttt{0} & {\scriptsize $\pm$\texttt{0}} & \texttt{0} & {\scriptsize $\pm$\texttt{0}} & \texttt{0} & {\scriptsize $\pm$\texttt{0}} & \texttt{0} & {\scriptsize $\pm$\texttt{0}} & \texttt{0} & {\scriptsize $\pm$\texttt{0}} & \texttt{0} & {\scriptsize $\pm$\texttt{0}} & \texttt{0} & {\scriptsize $\pm$\texttt{0}} & \texttt{0} & {\scriptsize $\pm$\texttt{0}} & \texttt{0} & {\scriptsize $\pm$\texttt{0}} & \texttt{8} & {\scriptsize $\pm$\texttt{5}} & \texttt{0} & {\scriptsize $\pm$\texttt{0}} & \textcolor{mygreen}{\texttt{13}} & {\scriptsize $\pm$\texttt{10}} \\
\texttt{humanoidmaze-giant-task2} & \texttt{0} & {\scriptsize $\pm$\texttt{0}} & \texttt{0} & {\scriptsize $\pm$\texttt{0}} & \texttt{0} & {\scriptsize $\pm$\texttt{0}} & \texttt{0} & {\scriptsize $\pm$\texttt{0}} & \texttt{0} & {\scriptsize $\pm$\texttt{0}} & \textcolor{mygreen}{\texttt{15}} & {\scriptsize $\pm$\texttt{6}} & \texttt{0} & {\scriptsize $\pm$\texttt{0}} & \texttt{0} & {\scriptsize $\pm$\texttt{0}} & \texttt{0} & {\scriptsize $\pm$\texttt{0}} & \texttt{40} & {\scriptsize $\pm$\texttt{18}} & \texttt{30} & {\scriptsize $\pm$\texttt{12}} & \texttt{25} & {\scriptsize $\pm$\texttt{21}} & \textcolor{mygreen}{\texttt{53}} & {\scriptsize $\pm$\texttt{17}} \\
\texttt{humanoidmaze-giant-task3} & \texttt{0} & {\scriptsize $\pm$\texttt{0}} & \texttt{0} & {\scriptsize $\pm$\texttt{0}} & \texttt{0} & {\scriptsize $\pm$\texttt{0}} & \texttt{0} & {\scriptsize $\pm$\texttt{0}} & \texttt{0} & {\scriptsize $\pm$\texttt{0}} & \texttt{0} & {\scriptsize $\pm$\texttt{0}} & \texttt{0} & {\scriptsize $\pm$\texttt{0}} & \texttt{0} & {\scriptsize $\pm$\texttt{0}} & \texttt{0} & {\scriptsize $\pm$\texttt{0}} & \texttt{0} & {\scriptsize $\pm$\texttt{0}} & \textcolor{mygreen}{\texttt{13}} & {\scriptsize $\pm$\texttt{5}} & \texttt{3} & {\scriptsize $\pm$\texttt{5}} & \texttt{5} & {\scriptsize $\pm$\texttt{6}} \\
\texttt{humanoidmaze-giant-task4} & \texttt{0} & {\scriptsize $\pm$\texttt{0}} & \texttt{0} & {\scriptsize $\pm$\texttt{0}} & \texttt{0} & {\scriptsize $\pm$\texttt{0}} & \texttt{0} & {\scriptsize $\pm$\texttt{0}} & \texttt{0} & {\scriptsize $\pm$\texttt{0}} & \texttt{0} & {\scriptsize $\pm$\texttt{0}} & \texttt{0} & {\scriptsize $\pm$\texttt{0}} & \texttt{0} & {\scriptsize $\pm$\texttt{0}} & \texttt{0} & {\scriptsize $\pm$\texttt{0}} & \texttt{15} & {\scriptsize $\pm$\texttt{10}} & \texttt{15} & {\scriptsize $\pm$\texttt{6}} & \texttt{10} & {\scriptsize $\pm$\texttt{8}} & \textcolor{mygreen}{\texttt{43}} & {\scriptsize $\pm$\texttt{22}} \\
\texttt{humanoidmaze-giant-task5} & \texttt{0} & {\scriptsize $\pm$\texttt{0}} & \texttt{0} & {\scriptsize $\pm$\texttt{0}} & \texttt{0} & {\scriptsize $\pm$\texttt{0}} & \textcolor{mygreen}{\texttt{3}} & {\scriptsize $\pm$\texttt{5}} & \textcolor{mygreen}{\texttt{3}} & {\scriptsize $\pm$\texttt{5}} & \textcolor{mygreen}{\texttt{3}} & {\scriptsize $\pm$\texttt{5}} & \texttt{0} & {\scriptsize $\pm$\texttt{0}} & \texttt{0} & {\scriptsize $\pm$\texttt{0}} & \texttt{0} & {\scriptsize $\pm$\texttt{0}} & \texttt{75} & {\scriptsize $\pm$\texttt{19}} & \texttt{58} & {\scriptsize $\pm$\texttt{22}} & \texttt{73} & {\scriptsize $\pm$\texttt{25}} & \textcolor{mygreen}{\texttt{88}} & {\scriptsize $\pm$\texttt{10}} \\
\addlinespace[0.2em]
\textit{Domain average} & \texttt{0} & {\scriptsize $\pm$\texttt{0}} & \texttt{0} & {\scriptsize $\pm$\texttt{0}} & \texttt{0} & {\scriptsize $\pm$\texttt{0}} & \texttt{1} & {\scriptsize $\pm$\texttt{1}} & \texttt{1} & {\scriptsize $\pm$\texttt{1}} & \textcolor{mygreen}{\texttt{4}} & {\scriptsize $\pm$\texttt{2}} & \texttt{0} & {\scriptsize $\pm$\texttt{0}} & \texttt{0} & {\scriptsize $\pm$\texttt{0}} & \texttt{0} & {\scriptsize $\pm$\texttt{0}} & \texttt{26} & {\scriptsize $\pm$\texttt{9}} & \texttt{25} & {\scriptsize $\pm$\texttt{5}} & \texttt{22} & {\scriptsize $\pm$\texttt{8}} & \textcolor{mygreen}{\texttt{40}} & {\scriptsize $\pm$\texttt{5}} \\
\midrule
\textit{Overall average} & \texttt{13} & {\scriptsize $\pm$\texttt{2}} & \texttt{34} & {\scriptsize $\pm$\texttt{3}} & \texttt{36} & {\scriptsize $\pm$\texttt{2}} & \texttt{29} & {\scriptsize $\pm$\texttt{2}} & \texttt{39} & {\scriptsize $\pm$\texttt{4}} & \texttt{39} & {\scriptsize $\pm$\texttt{4}} & \texttt{43} & {\scriptsize $\pm$\texttt{2}} & \textcolor{mygreen}{\texttt{48}} & {\scriptsize $\pm$\texttt{3}} & \texttt{14} & {\scriptsize $\pm$\texttt{2}} & \texttt{26} & {\scriptsize $\pm$\texttt{3}} & \texttt{27} & {\scriptsize $\pm$\texttt{4}} & \texttt{25} & {\scriptsize $\pm$\texttt{3}} & \textcolor{mygreen}{\texttt{43}} & {\scriptsize $\pm$\texttt{3}} \\
\bottomrule
\end{tabular}%
}
\endgroup
\end{table*}

\begin{table*}[t]
\centering
\caption{\footnotesize Complete results for explicit and implicit methods with $25\%$ of the dataset. We highlight nonzero mean performance at or above $95\%$ of the best in each row within each method group in {\color{mygreen}green}. ``\texttt{-}'' marks baselines we did not run on OCBench tasks. }
\label{tab:full-results-25}
\label{tab:full-results-explicit-25}
\label{tab:full-results-implicit-25}
\vspace{-0.5em}
\begingroup
\small
\setlength{\tabcolsep}{2pt}
\renewcommand{\arraystretch}{1.10}
\resizebox{\textwidth}{!}{%
\begin{tabular}{@{}l *{8}{r@{\hspace{0.05cm}}l}@{\hspace{0.18cm}} *{5}{r@{\hspace{0.05cm}}l}@{}}
\toprule
 & \multicolumn{16}{c}{\textit{Explicit methods}} & \multicolumn{10}{c}{\textit{Implicit methods}} \\
\cmidrule(lr){2-17}\cmidrule(l){18-27}
\texttt{Environment (OGBench)} & \hphantom{\texttt{00}}\makebox[0pt]{\texttt{Floq}} & \hphantom{\texttt{00}} & \hphantom{\texttt{00000}}\makebox[0pt]{\texttt{FlowCritic}} & \hphantom{\texttt{00000}} & \hphantom{\texttt{000}}\makebox[0pt]{\shortstack{\texttt{Value}\\\texttt{Flows}}} & \hphantom{\texttt{000}} & \hphantom{\texttt{00}}\makebox[0pt]{\texttt{PCBF}} & \hphantom{\texttt{00}} & \hphantom{\texttt{00}}\makebox[0pt]{\texttt{C51}} & \hphantom{\texttt{00}} & \hphantom{\texttt{00}}\makebox[0pt]{\texttt{IQN}} & \hphantom{\texttt{00}} & \hphantom{\texttt{0}}\makebox[0pt]{\texttt{QC}} & \hphantom{\texttt{0}} & \hphantom{\texttt{000}}\makebox[0pt]{\shortstack{\textcolor{mygreen}{\texttt{VF}}\\\textcolor{mygreen}{\texttt{(Ours)}}}} & \hphantom{\texttt{000}} & \hphantom{\texttt{00}}\makebox[0pt]{\texttt{DIVL}} & \hphantom{\texttt{00}} & \hphantom{\texttt{0000}}\makebox[0pt]{\shortstack{\texttt{Implicit}\\\texttt{C51}}} & \hphantom{\texttt{0000}} & \hphantom{\texttt{0000}}\makebox[0pt]{\shortstack{\texttt{Implicit}\\\texttt{IQN}}} & \hphantom{\texttt{0000}} & \hphantom{\texttt{00}}\makebox[0pt]{\texttt{Implicit QC}} & \hphantom{\texttt{00}} & \hphantom{\texttt{000}}\makebox[0pt]{\shortstack{\textcolor{mygreen}{\texttt{IVF}}\\\textcolor{mygreen}{\texttt{(Ours)}}}} & \hphantom{\texttt{000}} \\
\bottomrule
\end{tabular}%
}
\endgroup
\end{table*}

\begin{table*}[t]
\centering
\caption{\footnotesize Complete results for explicit and implicit methods with $50\%$ of the dataset. We highlight nonzero mean performance at or above $95\%$ of the best in each row within each method group in {\color{mygreen}green}.}
\label{tab:full-results-50}
\label{tab:full-results-explicit-50}
\label{tab:full-results-implicit-50}
\vspace{-0.5em}
\begingroup
\small
\setlength{\tabcolsep}{2pt}
\renewcommand{\arraystretch}{1.10}
\resizebox{\textwidth}{!}{%
\begin{tabular}{@{}l *{8}{r@{\hspace{0.05cm}}l}@{\hspace{0.18cm}} *{5}{r@{\hspace{0.05cm}}l}@{}}
\toprule
 & \multicolumn{16}{c}{\textit{Explicit methods}} & \multicolumn{10}{c}{\textit{Implicit methods}} \\
\cmidrule(lr){2-17}\cmidrule(l){18-27}
\texttt{Environment} & \hphantom{\texttt{00}}\makebox[0pt]{\texttt{Floq}} & \hphantom{\texttt{00}} & \hphantom{\texttt{00000}}\makebox[0pt]{\texttt{FlowCritic}} & \hphantom{\texttt{00000}} & \hphantom{\texttt{000}}\makebox[0pt]{\shortstack{\texttt{Value}\\\texttt{Flows}}} & \hphantom{\texttt{000}} & \hphantom{\texttt{00}}\makebox[0pt]{\texttt{PCBF}} & \hphantom{\texttt{00}} & \hphantom{\texttt{00}}\makebox[0pt]{\texttt{C51}} & \hphantom{\texttt{00}} & \hphantom{\texttt{00}}\makebox[0pt]{\texttt{IQN}} & \hphantom{\texttt{00}} & \hphantom{\texttt{0}}\makebox[0pt]{\texttt{QC}} & \hphantom{\texttt{0}} & \hphantom{\texttt{000}}\makebox[0pt]{\shortstack{\textcolor{mygreen}{\texttt{VF}}\\\textcolor{mygreen}{\texttt{(Ours)}}}} & \hphantom{\texttt{000}} & \hphantom{\texttt{00}}\makebox[0pt]{\texttt{DIVL}} & \hphantom{\texttt{00}} & \hphantom{\texttt{0000}}\makebox[0pt]{\shortstack{\texttt{Implicit}\\\texttt{C51}}} & \hphantom{\texttt{0000}} & \hphantom{\texttt{0000}}\makebox[0pt]{\shortstack{\texttt{Implicit}\\\texttt{IQN}}} & \hphantom{\texttt{0000}} & \hphantom{\texttt{00}}\makebox[0pt]{\texttt{Implicit QC}} & \hphantom{\texttt{00}} & \hphantom{\texttt{000}}\makebox[0pt]{\shortstack{\textcolor{mygreen}{\texttt{IVF}}\\\textcolor{mygreen}{\texttt{(Ours)}}}} & \hphantom{\texttt{000}} \\
\bottomrule
\end{tabular}%
}
\endgroup
\end{table*}

\begin{table*}[t]
\centering
\caption{\footnotesize Complete results for explicit and implicit methods with $100\%$ of the dataset. We highlight nonzero mean performance at or above $95\%$ of the best in each row within each method group in {\color{mygreen}green}. ``\texttt{-}'' marks baselines we did not run on OCBench tasks. }
\label{tab:full-results-100}
\label{tab:full-results-explicit-100}
\label{tab:full-results-implicit-100}
\vspace{-0.5em}
\begingroup
\small
\setlength{\tabcolsep}{2pt}
\renewcommand{\arraystretch}{1.10}
\resizebox{\textwidth}{!}{%
\begin{tabular}{@{}l *{8}{r@{\hspace{0.05cm}}l}@{\hspace{0.18cm}} *{5}{r@{\hspace{0.05cm}}l}@{}}
\toprule
 & \multicolumn{16}{c}{\textit{Explicit methods}} & \multicolumn{10}{c}{\textit{Implicit methods}} \\
\cmidrule(lr){2-17}\cmidrule(l){18-27}
\texttt{Environment (OGBench)} & \hphantom{\texttt{00}}\makebox[0pt]{\texttt{Floq}} & \hphantom{\texttt{00}} & \hphantom{\texttt{00000}}\makebox[0pt]{\texttt{FlowCritic}} & \hphantom{\texttt{00000}} & \hphantom{\texttt{000}}\makebox[0pt]{\shortstack{\texttt{Value}\\\texttt{Flows}}} & \hphantom{\texttt{000}} & \hphantom{\texttt{00}}\makebox[0pt]{\texttt{PCBF}} & \hphantom{\texttt{00}} & \hphantom{\texttt{00}}\makebox[0pt]{\texttt{C51}} & \hphantom{\texttt{00}} & \hphantom{\texttt{00}}\makebox[0pt]{\texttt{IQN}} & \hphantom{\texttt{00}} & \hphantom{\texttt{0}}\makebox[0pt]{\texttt{QC}} & \hphantom{\texttt{0}} & \hphantom{\texttt{000}}\makebox[0pt]{\shortstack{\textcolor{mygreen}{\texttt{VF}}\\\textcolor{mygreen}{\texttt{(Ours)}}}} & \hphantom{\texttt{000}} & \hphantom{\texttt{00}}\makebox[0pt]{\texttt{DIVL}} & \hphantom{\texttt{00}} & \hphantom{\texttt{0000}}\makebox[0pt]{\shortstack{\texttt{Implicit}\\\texttt{C51}}} & \hphantom{\texttt{0000}} & \hphantom{\texttt{0000}}\makebox[0pt]{\shortstack{\texttt{Implicit}\\\texttt{IQN}}} & \hphantom{\texttt{0000}} & \hphantom{\texttt{00}}\makebox[0pt]{\texttt{Implicit QC}} & \hphantom{\texttt{00}} & \hphantom{\texttt{000}}\makebox[0pt]{\shortstack{\textcolor{mygreen}{\texttt{IVF}}\\\textcolor{mygreen}{\texttt{(Ours)}}}} & \hphantom{\texttt{000}} \\
\bottomrule
\end{tabular}%
}
\endgroup
\end{table*}

\section{Experimental Details}
\label{appendix:exp-details}

\subsection{Dataset construction and evaluation protocol}
\label{appendix:data-and-eval-proto}

\textbf{OGBench datasets.} Following~\citet{zhou2026test}, we use OGBench datasets with $100$M transitions~\citep{park2024ogbench,park2025horizon} as the full offline dataset for \texttt{cube-triple}, \texttt{cube-quadruple}, \texttt{puzzle-4x4} and \texttt{humanoidmaze-giant}. These datasets are collected by scripted policies: \texttt{humanoidmaze-giant} has trajectories with $4000$ time steps, and the other datasets have trajectories with $1000$ time steps.

To subsample the full offline datasets with fractions within $\{10\%, 25\%, 50\%\}$, we draw a specific fraction of all trajectories in the offline dataset. Specifically, for each trajectory in the dataset, we compute the mean of rewards over time steps as the score and use a softmax distribution over these scores to sample trajectories without replacement. For each random seed, we sample the subset once before training and keep it fixed throughout training.

\textbf{OCBench datasets.} We additionally evaluate $6$ tasks from OCBench~\citep{park2026behavioral}: \texttt{block-lite-single-task1}, \texttt{block-lite-double-\{task1,task2\}}, and \texttt{chamber-lite-easy-\{task1,task2,task3\}}. These tasks use $7$-dimensional actions and operate at $10$ Hz. The \texttt{lite} variants have shorter horizons and simplified scripted policies compared to the original tasks~\citep{park2026behavioral}. We evaluate \methodname{}, \methodnamei{}, and selected baselines using $25\%$ and $100\%$ of the available offline data. The full training datasets contain $10{,}000$ episodes per task ($30{,}000$ for the two \texttt{block-lite-double} tasks), with $0.77$M transitions for \texttt{block-lite-single-task1}, $5.81$M and $4.79$M transitions for \texttt{block-lite-double-\{task1,task2\}}, and $2.11$M, $3.59$M and $3.78$M transitions for \texttt{chamber-lite-easy-\{task1,task2,task3\}}. The $25\%$ subsets are constructed by the same procedure as for OGBench datasets.

\textbf{Evaluation.} For evaluation, we follow a similar protocol as in~\citet{zhou2026test}. For each task in the OGBench benchmarks, the agent aims to reach a fixed goal. The episode terminates if the agent reaches the goal or exceeds the maximum number of time steps: $1000$ for \texttt{cube-triple} and \texttt{cube-quadruple}, $500$ for \texttt{puzzle-4x4}, and $4000$ for \texttt{humanoidmaze-giant}. For OCBench tasks, the episode limits are $200$ steps for \texttt{block-lite-single}, $400$ for \texttt{block-lite-double}, and $500$ for \texttt{chamber-lite-easy}, with early termination upon task completion. In Sec.~\ref{subsec:offline-exps} and Sec.~\ref{subsec:design-decisions}, we train all methods for $500$K gradient steps and, on OGBench, evaluate them every $100$K gradient steps for $10$ episodes per task. On OCBench, we follow the evaluation protocol of~\citet{park2026behavioral} and evaluate every $50$K gradient steps for $500$ episodes per task, running the episodes in parallel using OCBench's GPU-based simulator. On both benchmarks, we report the success rate at the final evaluation in Sec.~\ref{subsec:offline-exps}, showing the mean and one standard deviation across $4$ random seeds. In our ablation experiments (Sec.~\ref{subsec:design-decisions} and Appendix~\ref{appendix:hyperparam-ablations}), we report the success rate averaged over the last three evaluations: $300$K, $400$K, and $500$K gradient steps. For figures that aggregate results over different tasks or domains, we first average the success rate at the same random seed and then compute the mean and the standard deviation over different random seeds.

\subsection{Methods for comparison}
\label{appendix:prior-methods}

We compare VF and IVF against $11$ baselines in two categories spanning explicit methods and implicit methods (Sec.~\ref{sec:experiments}): $9$ prior RL methods and $2$ implicit variants of prior distributional RL methods that we construct for comparison. Below, we describe the formulation of each method.

\noindent\textbf{(1) Explicit methods.}

\paragraph{Floq~\citep{agrawalla2026floq}.}
Floq fits a scalar $Q$ function using flow matching. Let
$v_Q(z^\tau \mid s,a,\tau)$ denote its $Q$-velocity and
$\phi_Q(z^0 \mid s,a,\tau)$ denote the corresponding flow. Given a transition $(s, a, r, s')$, Floq constructs a TD target by averaging over $M$ bootstrapped samples:
\begin{align*}
    y &= r + \frac{\gamma}{M} \sum_{i=1}^{M}
    \bar{\phi}_Q(z_i^0 \mid s',a_i',1),
\end{align*}
where $z_i^0$ is sampled from the initial noise distribution $p_0(z_i^0) \overset{d}{=} \text{Unif}(\kappa [r_{\min}, r_{\max}] / (1-\gamma))$, where $\kappa$ is the noise coverage, $a_i'$ is the next action sampled from a one-step policy that transports Gaussian noise: $ a_i' = \pi_{\text{one-step}}(s', a_i^0)$, with $a_i^0 \sim \gN(0, I_{\abs{\gA}})$, and $\bar{\phi}_Q$ is the flow induced by the target $Q$-velocity $\bar{v}_Q$. Here, $\pi_{\text{one-step}}$ denotes a one-step flow policy as in~\citet{park2025flow}. Each TD target independently draws the initial return noise $z^0_i \sim p_0(z_i^0)$, and the policy noise $a^0_i$. To train the $Q$-velocity, Floq samples a noise level $\tau \sim \textsc{Unif}([0, 1])$, constructs the linear interpolant $z^\tau = (1 - \tau) z^0 + \tau y$, and fits
\begin{align*}
    \argmin_{v_Q} \: \E_{ \substack{ (s, a, r, s') \sim D, \: \tau \sim \textsc{Unif}([0, 1]), \\ z^0 \sim \textsc{Unif}( \kappa [r_{\text{min}}, r_{\text{max}}] / (1 - \gamma) ), \\ \left\{ a_i^0 \sim \gN(0, I_{\abs{\gA}}) \right\}_{i = 1}^M, \\ \left\{ z_i^0 \sim \textsc{Unif}(\kappa [r_{\text{min}}, r_{\text{max}}] / (1 - \gamma)) \right\}_{i = 1}^M } } \left[ (v_Q(z^\tau \mid s, a, \tau) - (y - z^0))^2 \right].
\end{align*}
This loss fits a velocity to transport initial noise toward the averaged TD
target. Floq therefore uses the flow to estimate the scalar $Q$ rather than directly fitting the return distribution. The mean flow endpoint is then distilled into a scalar $Q$-estimate,
$Q(s,a) \approx \E_{z^0 \sim p_0}[\phi_Q(z^0 \mid s,a,1)]$. As in~\citet{park2025flow}, Floq learns a one-step flow policy to maximize the critic estimate while staying close to a flow BC policy:
\begin{align*}
    \argmin_{\pi_{\text{one-step}}} \:
    \E_{s \sim D,\: a^0 \sim \gN(0, I)}
    \left[
        -Q(s,\pi_{\text{one-step}}(s, a^0))
        + \alpha
        \left\|
            \pi_{\text{one-step}}(s, a^0)
            - \pi_\beta(s,a^0)
        \right\|_2^2
    \right],
\end{align*}
where $\pi_\beta(s, a^0)$ denotes the action obtained by integrating
the BC flow from the same initial noise $a^0$, and $\alpha$ controls
the regularization strength.

\paragraph{FlowCritic~\citep{zhong2025flowcritic}.} FlowCritic models the full return distribution with flow matching. This method constructs TD targets using the return at the flow endpoint and then fits them using the standard flow-matching objective (Eq.~\ref{eq:standard-flow-matching-loss}). Specifically, let $v_Q(z^\tau \mid s, a, \tau)$ denote its $Q$-velocity and $p_Q(z^\tau \mid s, a, \tau)$ denote the corresponding probability density path. Given a transition $(s, a, r, s')$, we select the next action $a'$ from the policy $\pi(a' \mid s')$. FlowCritic constructs the TD target $y = r + \gamma z^1$, where $z^1 \sim \bar{p}_Q(z^1 \mid s', a', 1)$ is sampled from the probability density induced by the target $Q$-velocity $\bar{v}_Q$ at the next time step, and $a'$ is the next action sampled from the policy $\pi(a' \mid s')$ that is based on flow rejection sampling as in Eq.~\ref{eq:flow-rej-sampling}. To train the critic velocity, FlowCritic samples a noise level $\tau \sim \textsc{Unif}([0, 1])$, constructs the linear interpolant $z^\tau = (1 - \tau) z^0 + \tau y$, and fits
\begin{align}
    \argmin_{v_Q} \E_{ \substack{ (s, a, r, s') \sim D, \: a' \sim \pi(a' \mid s') \\ 
    \tau \sim \textsc{Unif}([0, 1]), \: z^0 \sim \gN(0, 1), \\ z^1 \sim \bar{p}_Q(z^1 \mid s', a', 1) }} \left[ (v_Q(z^\tau \mid s, a, \tau) - (y - z^0))^2 \right].
    \label{eq:flow-critic-loss}
\end{align}
Importantly, because the $Q$-velocity loss only uses the \emph{clean} return sample at the next time step, it does not enforce Bellman consistency at intermediate noise levels. For flow rejection sampling, FlowCritic generates $M$ MC return samples from the learned $Q$-velocity and discards the $m$ largest ones to construct the critic estimate:
\begin{align*}
    z_i^1 = z^0_i + \int_0^1 v_Q(z^\tau_i \mid s, a, \tau) d\tau, \: z^0_i \sim \gN(0, 1) \overset{\text{ranking}}{\implies} z_{1}^1 \leq \cdots \leq z_{M}^1, \quad  Q(s,a) = \frac{1}{M - m}\sum_{i=1}^{M - m} z_{i}^1.
\end{align*}

\paragraph{Value Flows~\citep{dong2025value}.} Value Flows models the return distribution using two complementary flow-matching objectives: DCFM loss and BCFM loss. The DCFM loss aims to incorporate the standard distributional Bellman equation (Eq.~\ref{eq:dist-bellman-eq}) into every noise level. However, as discussed in Sec.~\ref{subsec:flow-bellman-eq}, naively applying the standard distributional Bellman equation at all noise levels results in incorrect supervision at the noise level $\tau = 0$. Specifically, let $v_Q(z^\tau \mid s, a, \tau)$ denote the $Q$-velocity and
$p_Q(z^\tau \mid s, a, \tau)$ denote the corresponding probability density path. Given a transition
$(s,a,r,s') \sim D$, we sample the next action
$a' \sim \pi(a' \mid s')$ using flow rejection sampling
(Eq.~\ref{eq:flow-rej-sampling}). The DCFM loss
bootstraps directly from the target $Q$-velocity at the
next state: we generate a noisy return $z^{\tau} \sim \bar{p}_Q(z^\tau \mid s', a', \tau)$ and minimize
\begin{align*}
     \gL_{\text{DCFM}}(v_Q) = \E_{ \substack{(s, a, r, s') \sim D, \: a' \sim \pi(a' \mid s'), \\ \tau \sim \textsc{Unif}([0, 1]), \\ z^\tau \sim \bar{p}_Q(z^\tau \mid s', a', \tau)  }} \left[
        \left(
            v_Q(r+\gamma z^{\tau} \mid s,a,\tau)
            - \bar{v}_Q(z^{\tau} \mid s',a',\tau)
        \right)^2
    \right],
\end{align*}
where $\bar{v}_Q$ is the target $Q$-velocity. To stabilize learning, Value Flows also includes a BCFM loss $\gL_{\text{BCFM}}(v_Q)$, which is the same as the objective used by FlowCritic in Eq.~\ref{eq:flow-critic-loss}. Following the original implementation, we also include a confidence weight $w(s, a, z^0)$ with temperature $\lambda_w$ to reweight both $\gL_{\text{DCFM}}$ and $\gL_{\text{BCFM}}$. The final loss function is $\gL_{\text{DCFM}} + \lambda \gL_{\text{BCFM}}$ with $\lambda$ balancing the strengths of both losses. For flow rejection sampling, Value Flows selects action candidates using a critic estimate derived from one-step integration of return samples: $Q(s, a) \approx z^0 + v_Q(z^0 \mid s, a, 0), z^0 \sim \gN(0,1)$.

\paragraph{PCBF~\citep{xu2026path}.} PCBF models return distributions using flow matching with two return flow paths sharing the same initial noise. It also includes a control-variate target. Specifically, given a transition $(s, a, r, s') \sim D$, we sample the next action $a' \sim \pi(a' \mid s')$ using flow rejection sampling (Eq.~\ref{eq:flow-rej-sampling}). We draw $z^0 \sim \gN(0, 1)$ and generate a return at the next time step: $z' = \bar{\phi}_Q(z^0 \mid s', a', 1)$, where $\bar{\phi}_Q$ is the return flow path corresponding to the target velocity $\bar{v}_Q$. Using the same noise sample, PCBF constructs the two interpolation paths:
\begin{align*}
    \tilde{z}^{\tau} &= (1-\tau)z^0 + \tau z', \\
    z^{\tau} &= (1-\tau)z^0 + \tau(r+\gamma z')
    = \tau r + \gamma \tilde{z}^{\tau}
      + (1-\tau)(1-\gamma)z^0,
\end{align*}
where the second interpolation preserves the initial noise $z^0$ at $\tau = 0$, while the other endpoint satisfies $z^1 = r + \gamma z'$. To reduce the variance of the sample-based velocity target, PCBF introduces the control-variate correction
\begin{align*}
    u_{\lambda}
    = r+\gamma z'-z^0
    + \lambda \left[
        \bar{v}_Q(\tilde{z}^{\tau} \mid s',a',\tau)
        - (z'-z^0)
    \right],
\end{align*}
with $\lambda$ as a balancing coefficient, and minimizes
\begin{align*}
    \argmin_{v_Q} \E_{\substack{
        (s,a,r,s') \sim D,\: a' \sim \pi(a' \mid s'),\\
        z^0 \sim \gN(0,1),\:
        \tau \sim \textsc{Unif}([0,1])
    }}
    \left[
        \left(
            v_Q(z^{\tau} \mid s,a,\tau)-u_{\lambda}
        \right)^2
    \right],
\end{align*}
Setting $\lambda = 0$ removes the control-variate correction and leaves a flow-matching objective similar to FlowCritic.

\paragraph{C51~\citep{bellemare2017distributional}.}
C51 represents the return distribution as a categorical distribution over fixed atoms on the return support. This method minimizes the cross-entropy between this distribution and the projection of the distributional Bellman target onto the atoms. Following~\citet{dong2025value}, we pair C51 with a flow BC policy and rejection sampling (Eq.~\ref{eq:flow-rej-sampling}).

\paragraph{IQN~\citep{dabney2018implicit}.} IQN is a distributional RL method that learns the full return distribution by modeling the quantile function at randomly sampled quantile fractions. This method learns to fit the quantile function using a Huber loss with a threshold $\kappa$. Following~\citet{dong2025value}, we pair IQN with a flow BC policy and rejection sampling (Eq.~\ref{eq:flow-rej-sampling}).

\paragraph{QC~\citep{li2025reinforcement}.} Q-chunking (QC) learns a critic over action chunks with standard TD learning
and selects actions by rejection sampling using a flow BC policy (Eq.~\ref{eq:flow-rej-sampling}). 

\paragraph{\methodname{} (Ours).} \methodname{} follows the {\color{mygreen} pseudocode} in Alg.~\ref{alg:value-forcing}. We learn the $Q$-velocity and the $V$-velocity using the regression objectives derived from the corresponding stochastic processes (Eq.~\ref{eq:value-ret-fm-loss} and Eq.~\ref{eq:critic-ret-fm-loss}). We learn a flow BC policy and select actions using rejection sampling as in Eq.~\ref{eq:flow-rej-sampling}. We also use this rejection sampling to select the actions in the $V$-velocity loss (Eq.~\ref{eq:value-ret-fm-loss}). During training, \methodname{} also selects action candidates using a critic estimate derived from one-step integration of return samples: $Q(s, a) \approx z^0 + v_Q(z^0 \mid s, a, 0), z^0 \sim \gN(0,1)$.

\noindent\textbf{(2) Implicit methods.}

\paragraph{DIVL~\citep{wang2026learning}.} DIVL replaces scalar expectile regression with a categorical model of the state-action values associated with dataset actions. Let $p_{V, i}(s)$ denote the probability assigned to atom $c_i$ within a set of atoms $\{c_i\}_{i = 1}^{N}$, and let $\bar{Q}(s,a)$ denote the target $Q$ estimate. The categorical model minimizes
\begin{align*}
    -\E_{(s,a)\sim D}
    \left[
        \sum_{i=1}^{N}
        m_i\bigl(\bar{Q}(s,a)\bigr)
        \log p_{V, i}(s)
    \right],
\end{align*}
where $m_i(q)$ is the two-hot projection of $q$ onto the neighboring atoms. Thus, the learned model captures variation in state-action values across dataset actions at each state. DIVL extracts a bootstrap target from this model using an adaptive quantile:
\begin{align*}
    \bar{\mathcal{H}}(s)
    &= -\frac{1}{\log N}
       \sum_{i=1}^{N}p_{V,i}(s)\log p_{V,i}(s), \quad \varrho(s) = \operatorname{clip}
       \left(
           \varrho_{\mathrm{base}}
           -\eta\,\bar{\mathcal{H}}(s),\,0,\,1
       \right), \\
    V_{\varrho}(s)
    &= \min\left\{
        c_i:\sum_{j=1}^{i}p_{V,j}(s)\geq\varrho(s)
       \right\}.
\end{align*}
A higher entropy $\bar{H}(s)$ lowers the selected quantile, making the bootstrap target less optimistic. For a transition $(s, a, r, s') \sim D$, the $Q$ estimate minimizes
\begin{align*}
    \E_{(s, a, r, s')\sim D}
    \left[
        \left(
            Q(s, a)
            -\left(r + \gamma V_{\varrho}(s')\right)
        \right)^2
    \right].
\end{align*}
We replace the original adjoint-matching policy extraction with flow rejection sampling (Eq.~\ref{eq:flow-rej-sampling}), ranking action candidates by $Q(s, a)$. 

\paragraph{Implicit C51.} Implicit C51 is an implicit variant of C51~\citep{bellemare2017distributional} that we construct for comparison. Instead of using the expectile weight to fit the $V$-velocity as in Eq.~\ref{eq:value-ret-imp-vel-loss}, we use the expectile weight to fit the $V$-return distribution. Let $p_{Q, i}(s,a)$ and $p_{V, i}(s)$ denote the probabilities assigned to atom $c_i$ by the $Q$-model and the $V$-model, respectively. Their scalar means are
\begin{align}
    Q(s,a) &= \sum_{i = 1}^{N} c_i p_{Q, i}(s,a),
    &
    V(s) &= \sum_{i = 1}^{N} c_i p_{V, i}(s).
    \label{eq:ic51-q-v}
\end{align}
For each transition $(s, a, r, s')$, we minimize the expectile weighted cross-entropy
\begin{align*}
    -\E_{(s,a)\sim D}
       \left[
           w^\mu(s,a)
           \sum_{i=1}^{N}
           \bar{p}_{Q,i}(s,a)
           \log p_{V,i}(s)
       \right].
\end{align*}
with $w^\mu(s, a)$ as in Eq.~\ref{eq:expectile-weight} but defined by the $Q$ and $V$ in Eq.~\ref{eq:ic51-q-v}, to learn the $V$-return distribution. For the $Q$-return distribution, we use the $V$-return distribution at the next state $s'$ to construct the TD target.

\paragraph{Implicit IQN.} Implicit IQN is an implicit variant of IQN that we construct for comparison. Following~\citet{kostrikov2022offline}, implicit IQN adds a $V$-quantile $q_V(s, \omega)$ that uses the IQN network architecture without the action input. Let $q_Q(s, a, \omega)$ denote the $Q$-quantile at quantile fraction $\omega \sim \textsc{Unif}([0,1])$. The scalar $Q$ and $V$ estimates can be written as 
\begin{align}
    Q(s, a)= \E_{\omega}[q_Q(s, a, \omega)], \quad V(s) = \E_{\omega}[q_V(s, \omega)].
    \label{eq:iiqn-q-v}
\end{align}
Similar to Implicit C51, we learn the $V$-quantile by minimizing the expectile-weighted loss
\begin{align*}
    \E_{\substack{
        (s,a) \sim D, \: \omega \sim \textsc{Unif}([0, 1])
    }}
    \left[
        w^\mu(s,a)
        \left(
            q_V(s, \omega) - \bar{q}_Q(s, a, \omega)
        \right)^2
    \right],
\end{align*}
with $w^\mu(s, a)$ as in Eq.~\ref{eq:expectile-weight} but defined by the $Q$ and $V$ in Eq.~\ref{eq:iiqn-q-v}. For the $Q$-quantile, we use the $V$-quantile at the next state $s'$ to construct the TD target for the quantile Huber loss with threshold $\kappa$:
\begin{align*}
    \rho_{\omega}^{\kappa}(u)
    &= \abs{ \omega - \mathbbm{1} \{ u < 0 \} }
       \frac{\mathcal{H}_{\kappa}(u)}{\kappa},
    &
    \mathcal{H}_{\kappa}(u)
    &=
    \begin{cases}
        \frac{1}{2}u^2, & |u|\leq\kappa,\\
        \kappa\left(|u|-\frac{1}{2}\kappa\right),
        & |u|>\kappa.
    \end{cases}
\end{align*}

\paragraph{Implicit QC~\citep{hansen2023idql, li2025reinforcement}.} Implicit QC combines the scalar $Q$ and $V$ estimates from IQL~\citep{kostrikov2022offline} with best-of-$N$ rejection sampling from a flow BC policy. To make it consistent with other methods, we use $K$ to denote the number of action candidates instead of $N$. Unlike QC, which constructs TD targets using the action chunk sampled from the policy, Implicit QC learns the $V$-function using expectile regression and the $Q$-function using TD targets based on the next-state $V$ estimate.

\paragraph{IVF (Ours).} \methodnamei{} follows the {\color{mygreendark} pseudocode} in Alg.~\ref{alg:value-forcing}. Similar to \methodname{}, we learn the $Q$-velocity and the $V$-velocity using the regression objectives derived from the corresponding stochastic processes (Eq.~\ref{eq:value-ret-imp-vel-loss} and Eq.~\ref{eq:critic-ret-fm-loss}). We learn a flow BC policy and select actions using rejection sampling as in Eq.~\ref{eq:flow-rej-sampling}.

\subsection{Implementations and hyperparameters}
\label{apdx:implementation-and-hyperparameters}

\begin{table}[t]
\centering
\caption{Default hyperparameters for \methodname{}, \methodnamei{} and all baselines.}
\label{tab:common-hyperparams}
\begin{tabular}{@{}>{\raggedright\arraybackslash}p{0.31\linewidth}>{\raggedright\arraybackslash}p{0.64\linewidth}@{}}
\toprule
Hyperparameter & Value \\
\midrule
Optimizer & Adam~\citep{kingma2014adam} \\
Learning rate & $3\times10^{-4}$ \\
Gradient steps & 500K \\
Minibatch size & 1024 \\
Network architecture & OGBench: MLP with hidden dimensions $[1024,1024,1024,1024]$; OCBench: residual MLP with an input layer and $4$ two-layer residual blocks of width $1024$ \\
Nonlinearity & GELU~\citep{hendrycks2016gaussian} \\
Layer normalization & Yes \\
Noise-level embedding & 16-dimensional sinusoidal \\
Noise-level distribution & $\textsc{Unif}([0, 1])$ \\
Target network update rate & $0.005$ \\
Discount factor $\gamma$ & OGBench: $0.999$; OCBench: $0.9999$  \\
Action chunk length $H$ & $5$ \\
BC flow integration steps & $10$ (\methodname{} = 4 during training) \\
Action candidates $K$ & $16$ \\
Return-flow integration steps & 10 \\
\bottomrule
\end{tabular}
\end{table}

\begin{table}[t]
\centering
\caption{Domain-specific hyperparameters.}
\label{tab:domain-specific-hyperparams}
\small
\setlength{\tabcolsep}{4pt}
\begin{tabular}{@{}lccccccc@{}}
\toprule
& \multicolumn{7}{c}{\textit{Explicit methods}} \\
\cmidrule(l){2-8}
& \multicolumn{3}{c}{Floq} & \multicolumn{2}{c}{Value Flows} & PCBF & IQN \\
\cmidrule(lr){2-4}\cmidrule(lr){5-6}\cmidrule(lr){7-7}\cmidrule(l){8-8}
Domain & $\alpha$ & \texttt{vel-depth} & $\kappa$ & $\lambda$ & $\lambda_w$ & $\lambda$ & $\kappa$ \\
\midrule
\texttt{cube-triple}        & 300  & 2 & 0.1  & 0.3 & 0.3 & 0.6 & 300 \\
\texttt{cube-quadruple}     & 300  & 2 & 0.1  & 0.3 & 0.3 & 0.6 & 300 \\
\texttt{puzzle-4x4}         & 3000 & 4 & 0.25 & 3   & 1   & 0.4 & 300 \\
\texttt{humanoidmaze-giant} & 300  & 4 & 0.1  & 3   & 0.3 & 0.2 & 30  \\
\texttt{OCBench (all tasks)} & \texttt{-} & \texttt{-} & \texttt{-} & 0.3 & 0.3 & \texttt{-}  & \texttt{-} \\
\bottomrule
\end{tabular}

\vspace{0.8em}
\begin{tabular}{@{}lccccc@{}}
\toprule
& & \multicolumn{4}{c}{\textit{Implicit methods}} \\
\cmidrule(lr){2-2}\cmidrule(l){3-6}
& All methods & DIVL & Implicit C51 & Implicit IQN & \methodnamei{} \\
Domain & $[r_{\min},r_{\max}]$ & $\varrho_{\mathrm{base}}$ & $\mu$ & $\kappa$ & $\mu$ \\
\midrule
\texttt{cube-triple}         & $[-3,0]$  & 0.9 & 0.9 & 300 & 0.9 \\
\texttt{cube-quadruple}      & $[-4,0]$  & 0.9 & 0.8 & 300 & 0.8 \\
\texttt{puzzle-4x4}          & $[-16,0]$ & 0.9 & 0.9 & 300 & 0.9 \\
\texttt{humanoidmaze-giant}  & $[-1,0]$  & 0.9 & 0.8 & 30  & 0.8 \\
\texttt{OCBench (all tasks)} & $[-1,0]$  & \texttt{-} & 0.9 & \texttt{-} & 0.9 \\
\bottomrule
\end{tabular}
\end{table}

We implement \methodname{} and \methodnamei{} and all baselines in a~\texttt{JAX}~\citep{jax_autodiff_cookbook} codebase, adapting implementations from Floq~\citep{agrawalla2026floq}, Value Flows~\citep{dong2025value}, PCBF~\citep{xu2026path}, FQL~\citep{park2025flow}, and QGF~\citep{zhou2026test}. We describe the implementation details and hyperparameters for each method below. 

\paragraph{Hyperparameter tuning.} Following \citet{zhou2026test}, we tune method-specific hyperparameters per domain rather than per task: we train on tasks $2$ and $4$ of each domain with the full dataset, and average the success rates at 500K gradient steps
over tasks and seeds. We then use the selected values for all $5$ tasks and all data budgets. We show common hyperparameters in Table~\ref{tab:common-hyperparams} and domain-specific hyperparameters in Table~\ref{tab:domain-specific-hyperparams}.

\noindent\textbf{(1) Explicit methods.}

\paragraph{Floq.} We use $\alpha$ to denote the BC regularization coefficient. Following their original implementation, the $Q$-velocity preprocesses the noisy critic using an HL-Gauss histogram~\citep{imani2018improving} and preprocesses the noise level using cosine embedding. We use a uniform distribution $\textsc{Unif}(\kappa \cdot [ r_{\text{min}}, r_{\text{max}}] / (1 - \gamma))$ to sample the initial noise. We use an MLP with $512$ units and different depths (\texttt{vel-depth}) to parameterize the $Q$-velocity. For $\kappa$ and $\texttt{vel-depth}$, we reuse values from the most similar domains in the original implementations. We sweep $\alpha$ within $\alpha\in\{300, 1000, 3000, 10000\}$ on each domain separately.

\paragraph{FlowCritic.} To speed up training, flow rejection sampling uses $M_\text{train} = 4$ with $1$ Euler step (Tweedie's formula~\citep{dong2025value, zhou2026test}). During evaluation, we use $M_\text{eval} = 10$ with $10$ Euler steps. For both phases, we discard the largest return sample ($m = 1$). We do not tune hyperparameters for FlowCritic.

\paragraph{Value Flows.} Value Flows learns the return distribution with a two-member $Q$-velocity similar to the twin Q trick in~\citet{fujimoto2018addressing}.
Following the original implementation, we clip returns to the range $[r_{\text{min}},r_{\text{max}}] / (1-\gamma)$ after every integration step of the return flow. We also clip the BC actions after every Euler step. We tune $\lambda \in\{0.3,1,3,10\}$ and $\lambda_w \in\{0.3,1\}$ jointly.

\paragraph{PCBF.} PCBF also learns the return distribution with a two-member $Q$-velocity similar to the twin Q trick in~\citet{fujimoto2018addressing}. We also use the same return flow clipping and action flow clipping as in Value Flows. We sweep $\lambda \in \{0.3,0.4,0.6\}$ on the two \texttt{cube} domains, $\lambda \in \{0.1, 0.2, 0.4\}$ on \texttt{puzzle-4x4}, and fix $\lambda = 0.2$ on \texttt{humanoidmaze-giant}.

\paragraph{C51.} We use the twin Q trick from~\citet{fujimoto2018addressing} in C51. We use a fixed number of $51$ atoms spread evenly on $[r_{\min},r_{\max}]/(1-\gamma)$. We do not tune C51.

\paragraph{IQN.} IQN uses a network that takes in a quantile fraction within $(0, 1)$ as input, and embeds it using $64$-dimensional cosine features. We also use the twin Q trick from~\citet{fujimoto2018addressing} in IQN. For estimating the $Q(s, a)$, we average the quantile predictions from $8$ random fractions, drawn independently from a uniform distribution $\textsc{Unif}([0, 1])$. The $(s, a)$ input and the
cosine features of the quantile are each projected linearly to $d_{\text{emb}} = 32$ dimensions. The two projections are multiplied elementwise and then passed to an MLP. We tune the threshold $\kappa$ within $\{ 30, 300, 3000 \}$.

\paragraph{QC.} We use the best-of-$N$ variant of QC. In this variant, the best of the $K$ action candidates at the next state is used as the next action in the TD target: the critic evaluates the rejection-sampling policy. We use the twin Q trick from~\citet{fujimoto2018addressing} in QC. We use $K = 16$ action candidates on every domain, as in the other methods. We do not tune QC. 

\paragraph{\methodname{} (Ours).} The $Q$-velocity and the $V$-velocity are each ensembles of two members as in the twin Q trick from~\citep{fujimoto2018addressing}. We use $K = 16$ action candidates during both training and evaluation. During training, we use $M = 1$ one-step integration of return samples to approximate the critic. During evaluation, \methodname{} ranks $K = 16$ action candidates by a full $10$-step integration over $M = 16$
return samples whose initial noise is shared across candidates to reduce variance. We do not tune hyperparameters for \methodname{}.

\noindent\textbf{(2) Implicit methods.}

\paragraph{DIVL.} We tune $\varrho_{\mathrm{base}}\in\{0.7,0.8,0.9\}$, while retaining $\eta=0.3$. We use $2001$ atoms on $[r_{\min}/(1-\gamma),r_{\max}/(1-\gamma)]$, increasing the number of atoms from the original $201$ to obtain finer quantile resolution.

\paragraph{Implicit C51.} We use the same atoms, support, and $Q$-model as C51. For the $V$-model, we use the same architecture as the $Q$-model. We adopt \methodnamei{}'s domain-specific expectile $\mu$ (Table~\ref{tab:domain-specific-hyperparams}) without separately tuning it for Implicit C51. 

\paragraph{Implicit IQN.} We reuse $\kappa$ from IQN and use $\mu = 0.9$ on all domains. We use $8$ sampled quantile fractions, $64$ cosine features, embedding dimension $d_{\text{emb}} = 32$. 

\paragraph{Implicit QC.} We use the twin Q trick from~\citep{fujimoto2018addressing} and an expectile of $\mu = 0.9$ to learn the V function. The Q networks are trained using the TD targets with the V networks. We do not tune hyperparameters for Implicit QC.

\paragraph{\methodnamei{} (Ours).} We use one network for both the $V$-velocity and $Q$-velocity as adapted from the classifier-free guidance literature~\citep{ho2022classifier}. This network contains two members as in the twin Q trick from~\citep{fujimoto2018addressing}. We tune $\mu \in \{0.7, 0.8, 0.9\}$ and fix $M = 16$ on all domains.

\subsection{\methodexplain{} improves return estimation accuracy}
\label{appendix:ret-hist}

For each domain, we restore a single state from the OGBench $100$M dataset, execute a fixed starting action chunk, and then follow a fixed data-collection policy until success or the time limit. The data-collection policy is the scripted policy with action noise or a noisy expert policy from~\citet{park2024ogbench}. From each representative state, we collect a training set and a held-out evaluation set, each with $1000$ trajectories. We use reward functions of \texttt{cube-triple-task5}, \texttt{cube-quadruple-task2}, \texttt{puzzle-4x4-task4} and
\texttt{humanoidmaze-giant-task2} to compute the return distributions.

We train \methodname{}, Value Flows, FlowCritic, PCBF, and \texttt{Oracle} on each training set for
$100$K gradient steps using each method's default hyperparameters on each domain. To evaluate the data-collection policy, all methods except \texttt{Oracle} bootstrap with the action chunk in the dataset (SARSA) on the training trajectories. For \texttt{Oracle}, it instead fits the discounted returns from different trajectories with the standard flow-matching loss
(Eq.~\ref{eq:standard-flow-matching-loss}). We draw $5000$ return samples from each estimator at the single state and action chunk pair to visualize the predicted return histograms and compute the $1$-Wasserstein distance to the $1000$ evaluation returns. Our histograms use $200$ bins between the $10$th percentile and the maximum of the ground-truth returns, widened to include all Value Flows samples. 

For the implicit methods in Fig.~\ref{fig:implicit-ret-hist}, we additionally evaluate IVF, Implicit C51, Implicit IQN, and DIVL using the same training trajectories, held-out evaluation returns, and evaluation protocol described above.

\subsection{Key design decisions in \methodexplain{}}
\label{appendix:alternative-algos}

Below, we describe the alternative instantiations of RL algorithms evaluated in Fig.~\ref{fig:value-policy-ablations}. Since we use a target $Q$-velocity $v^{\bar{\psi}}_Q$ to update the $V$-velocity in Alg.~\ref{alg:value-forcing}, we will use $\bar{v}_Q$ to simplify its notation and use $\bar{p}_Q$ to denote its induced noisy return distribution.

\paragraph{\methodname{} without the $V$-velocity.} Similar to prior actor-critic algorithms~\citep{fujimoto2018addressing, haarnoja2018soft} that only learn a $Q$-estimate, one variant of \methodname{} can only learn a $Q$-velocity by bootstrapping the exponential moving average of itself (\texttt{no $V$-velocity}).

Specifically, we can remove the $V$-velocity in \methodname{} and use the target $Q$-velocity at the next action $a' \sim \pi(a' \mid s')$ to sample the noisy return $z^\tau \sim \bar{p}_Q(z^\tau \mid s', a', \tau)$. We also replace the $V$-velocity $v_V$ in the regression target with the target $Q$-velocity $\bar{v}_Q$, resulting in the following objective for \texttt{no value velocity}:
\begin{align*}
    &\argmin_{v_Q} \: \E_{ \substack{s' \sim p(s' \mid s, a), \, a' \sim \pi(a' \mid s') \\
    \: \tau \sim \textsc{Unif}([0, 1]), \\
    \epsilon \sim \gN(0, 1), \: z^\tau \sim \bar{p}_Q(z^{\tau} \mid s', a', \tau )} } \left[ \left( v_Q \left( \tau(1 - \gamma) r + \gamma z^\tau + (1 - \tau) \sqrt{1 - \gamma^2} \epsilon \, \middle| \, s, a, \tau \right) \right. \right. \nonumber \\
    &\hspace{15.5em} \left. \left. - \left( (1 - \gamma) r + \gamma \bar{v}_Q(z^\tau \mid s', a', \tau) - \sqrt{1 - \gamma^2} \epsilon \right) \right)^2 \right],
\end{align*}
where the policy $\pi$ is the rejection sampling policy as in Eq.~\ref{eq:flow-rej-sampling}. This loss directly instantiates the flow Bellman backup without using the $V$-velocity in the critic loss.

\paragraph{\methodname{} with a one-step actor.}
Following~\citet{park2025flow}, the \texttt{one-step actor} variant of \methodname{} replaces rejection sampling with a one-step policy $\pi_{\text{one-step}}(s, a^0)$, where $a^0 \sim \gN(0, I_{\abs{\gA}})$. Let $\pi_\beta(s, a^0)$ denote the action generated by the BC flow policy from integrating the same initial noise $a^0$. We train the BC flow policy using the standard conditional flow matching loss $\gL(\pi_\beta)$ as in Eq.~\ref{eq:standard-flow-matching-loss}. Then, the one-step policy learns to maximize the scalar $Q$ estimate induced by our $Q$-velocity while distilling from the BC flow policy:
\begin{align}
    \argmin_{\pi_{\text{one-step}}} \: 
    \E_{s \sim \gD, \, a^0 \sim \gN(0, I_{\abs{\gA}})}
    \left[ -Q(s, \pi_{\text{one-step}}(s, a^0) )
    + \alpha
        \norm*{
            \pi_\text{one-step}(s, a^0)
            - \pi_\beta(s, a^0)
        }_2^2
    \right],
    \label{eq:one-step-actor-loss}
\end{align}
where $\alpha \geq 0$ controls the strength of the distillation regularizer. Again, we compute the $Q$ estimate by averaging over $M$ MC return samples and propagate gradients for the one-step actor through the return flow integration. For the ablation experiments, we tune $\alpha \in \{1, 3, 10 \}$ for each domain by averaging the success rates across task $2$ and task $4$ and choose $\alpha = 3$ for both \texttt{cube-triple} and \texttt{puzzle-4x4}. In addition, we use the one-step actor $\pi_\text{one-step}$ to sample actions in the $V$-velocity loss (Eq.~\ref{eq:value-ret-fm-loss}) during training and select actions during evaluation.

\paragraph{\methodnamei{} with naive expectile regression.} The most naive way to extend the expectile regression in~\citet{kostrikov2022offline} to return estimation is to directly apply the expectile loss to the velocity residual (as opposed to Eq.~\ref{eq:value-ret-imp-vel-loss}). We call this variant of IVF \texttt{expectile regression}. Specifically, we can define the naive expectile weight as
\begin{footalign*}
    w^\mu_\text{naive}(s, a, z^\tau, \tau) \triangleq \mu \mathbbm{1}( \bar{v}_Q(z^\tau \mid s, a, \tau) \geq v_V(z^\tau \mid s, \tau) ) + (1 - \mu) \mathbbm{1}( \bar{v}_Q(z^\tau \mid s, a, \tau) < v_V(z^\tau \mid s, \tau) ),
\end{footalign*}
and then use it to reweight the $V$-velocity loss:
\begin{align*}
    &\argmin_{v_V} \: \E_{ \substack{ a \sim D, \: \tau \sim \textsc{Unif}([0, 1]), \\ z^\tau \sim \bar{p}_Q(z^\tau \mid s, a, \tau) } } \left[ w^\mu_{\text{naive}}(s, a, z^\tau, \tau) ( \bar{v}_Q(z^\tau \mid s, a, \tau) - v_V(z^{\tau} \mid s, \tau) )^2 \right] \\
    &= \argmin_{v_V} \: \E_{ \substack{ a \sim D, \: \tau \sim \textsc{Unif}([0, 1]), \\ z^\tau \sim \bar{p}_Q(z^\tau \mid s, a, \tau) } } \left[ \ell^2_\mu \left( \bar{v}_Q(z^\tau \mid s, a, \tau) - v_V(z^{\tau} \mid s, \tau) \right) \right],
\end{align*}
where $\ell_\mu^2(\cdot)$ is the standard expectile regression loss with expectile $\mu$. However, when $\mu$ approaches $1$, applying Lemma 1 of~\citet{kostrikov2022offline}, the minimizer of the expectile regression loss above $v^{\mu, \: \text{naive}}_V$ satisfies 
\begin{align*}
    \lim_{\mu \to 1} v^{\mu, \: \text{naive}}_V(z^\tau \mid s, \tau) = \max_{a: \: \pi_{\beta}(a \mid s) > 0} \bar{v}_Q(z^\tau \mid s, a, \tau).
\end{align*}
Taking a pointwise expectile of velocity directly does not generally produce an expectile of the induced expected returns, since the velocity fields generate different trajectories. Integrating both sides gives an \emph{upper envelope} over the $Q$-velocity instead of the maximum Q estimate within the dataset support:
\begin{align*}
    \lim_{\mu \to 1} \E_{z^0 \sim \gN(0, 1)} \left[ \int_0^1 v^{\mu, \: \text{naive}}_V(z^\tau \mid s, \tau) d\tau \right] &= \E_{z^0 \sim \gN(0, 1)} \left[ \int_0^1 \max_{a: \: \pi_\beta(a \mid s) > 0} \bar{v}_Q(z^\tau \mid s, a, \tau) d\tau \right] \\
    &\neq \max_{a: \: \pi_\beta(a \mid s) > 0} \E_{z^0 \sim \gN(0, 1)} \left[ \int_0^1 \bar{v}_Q(z^\tau \mid s, a, \tau) d\tau \right] \\
    &= \max_{a: \: \pi_\beta(a \mid s) > 0} \bar{Q}(s, a).
\end{align*}
Thus, naively applying expectile regression to the return velocity residual does not recover the value and critic estimates as in~\citet{kostrikov2022offline}.

\paragraph{\methodnamei{} with quantile weights.} Prior work has found that a $1$-dimensional flow avoids crossing along the flow paths (Lemma 3.9 of~\citet{liu2022flow}). This property preserves quantiles (Lemma 7 of~\citep{bansal2026on}): the $q$-quantile of the noise distribution maps exactly to the $q$-quantile of the target distribution. Using this property, we can derive a variant of \methodnamei{} called \texttt{quantile weight} that uses the $q$-quantile of the return flows to construct the weights in Eq.~\ref{eq:value-ret-imp-vel-loss}. Specifically, for a quantile $q \in (0, 1)$, we can define the quantile weight with $\mu \in [0.5, 1)$ as
\begin{align*}
    w^\mu_q(s, a) \triangleq \mu \mathbbm{1}( \bar{Q}_q(s, a) \geq V_q(s) ) + (1 - \mu) \mathbbm{1} ( \bar{Q}_q(s, a) < V_q(s)),
\end{align*}
where $\bar{Q}_q(s, a)$ and $V_q(s)$ are $q$-quantiles of the $Q$-flow and the $V$-flow obtained by integrating the $q$-quantile of the standard Gaussian:
\begin{align*}
    \bar{Q}_q(s, a) = z^0 + \int_0^1 \bar{v}_Q(z^\tau \mid s, a, \tau) d\tau, \: V_q(s) = z^0 + \int_0^1 v_V(z^\tau \mid s, \tau) d\tau, \: z^0 = \text{quantile}_{\gN(0, 1)}(q).
\end{align*}
Note that we do not need to compute quantiles $\bar{Q}_q$ and $V_q$ using MC return samples because of the quantile-preserving property. For the ablation experiments, we tune $q \in \{0.7, 0.8, 0.9 \}$ for each domain by averaging the success rates across task $2$ and task $4$ and choose $q = 0.9$ for both \texttt{cube-triple} and \texttt{puzzle-4x4}.

\end{document}